%% file: final.tex
\documentclass[journal]{IEEEtran}
\usepackage{amsmath,amsfonts}
\usepackage{url}
\usepackage{marvosym}
\usepackage{balance}
\usepackage[hidelinks,hypertexnames=false]{hyperref}
\usepackage{graphicx}
\usepackage{cite}
\usepackage{booktabs}
\usepackage{multirow}
\usepackage{threeparttable}
\usepackage{makecell}
\usepackage{mathrsfs}
\usepackage{enumitem}
\usepackage{ulem}
\usepackage{color}
\usepackage{float}
\usepackage[switch]{lineno}
\input{packages.tex}
\newcolumntype{L}[1]{>{\raggedright\arraybackslash}p{#1}}
\newcolumntype{C}[1]{>{\centering\arraybackslash}p{#1}}
\newcolumntype{R}[1]{>{\raggedleft\arraybackslash}p{#1}}

\newcommand{\settablefont}{\fontsize{6.9}{11.8}\selectfont}

\begin{document}
\title{TiCoSS: Tightening the Coupling between\\Semantic Segmentation and Stereo Matching\\within A Joint Learning Framework}
\normalem
\author{
Guanfeng Tang$^{\orcidicon{0009-0002-5918-4775}\,}$, Zhiyuan Wu$^{\orcidicon{0009-0001-7253-7603}\,}$, Jiahang Li$^{\orcidicon{0009-0005-8379-249X}}$, Ping Zhong$^{\orcidicon{0000-0003-3393-8874}}$, \\ Wei Ye$^{\orcidicon{0000-0002-3784-7788}}$, Xieyuanli Chen$^{\orcidicon{0000-0003-0955-6681}}$, Huimin Lu$^{\orcidicon{0000-0002-6375-581X}}$
and Rui Fan$^{\orcidicon{0000-0003-2593-6596}\,}$,~\IEEEmembership{Senior Member,~IEEE}
\vspace{-1em}
\thanks{This research was supported in part by the National Natural Science Foundation of China under Grant 62473288, Grant 62233013, Grant 62403478, Grant 62176184, and Grant 62272489; in part by the Fundamental Research Funds for the Central Universities; in part by the NIO University Programme (NIO UP); in part by the Xiaomi Young Talents Program; and in part by the Young Elite Scientists Sponsorship Program by CAST (No. 2023QNRC001) (\emph{Corresponding author: Rui Fan}).}
\thanks{Guanfeng Tang and Jiahang Li are with the College of Electronics and Information Engineering, Tongji University, Shanghai 201804, China (e-mails: \{gftang, lijiahang617\}@tongji.edu.cn).}
\thanks{Wei Ye is with the College of Electronics and Information Engineering, Tongji University, Shanghai 201804, China. He is also with the Shanghai Innovation Institute, Shanghai 200231, China (e-mail: yew@tongji.edu.cn).}
\thanks{Zhiyuan Wu is with the Department of Engineering, King’s College London, London WC2R 2LS, United Kingdom (e-mail: zhiyuan.1.wu@kcl.ac.uk).}
\thanks{Ping Zhong is with the Department of Computer Science and Technology, Central South University. (email: ping.zhong@csu.edu.cn).}
\thanks{Xieyuanli Chen and Huimin Lu are with the College of Intelligence Science and Technology, National University of Defense Technology, Changsha 410082, China (e-mail: chenxieyuanli@hotmail.com; lhmnew@nudt.edu.cn).}
\thanks{
Rui Fan is with the College of Electronics and Information Engineering, Shanghai Institute of Intelligent Science and Technology, Shanghai Research Institute for Intelligent Autonomous Systems,  Shanghai Key Laboratory of Intelligent Autonomous Systems, State Key Laboratory of Autonomous Intelligent Unmanned Systems, and Frontiers Science Center for Intelligent Autonomous Systems of the Ministry of Education, Tongji University, Shanghai 201804, China (e-mail: rui.fan@ieee.org).}}

\maketitle

\begin{abstract}
Semantic segmentation and stereo matching, respectively analogous to the ventral and dorsal streams in our human brain, are two key components of autonomous driving perception systems. Addressing these two tasks with separate networks is no longer the mainstream direction in developing computer vision algorithms, particularly with the recent advances in large vision models and embodied artificial intelligence. The trend is shifting towards combining them within a joint learning framework, especially emphasizing feature sharing between the two tasks. The major contributions of this study lie in comprehensively tightening the coupling between semantic segmentation and stereo matching. Specifically, this study makes three key contributions: (1) a tightly coupled, gated feature fusion strategy, (2) a hierarchical deep supervision strategy, and (3) a coupling tightening loss function. The combined use of these technical contributions results in TiCoSS, a state-of-the-art joint learning framework that simultaneously tackles semantic segmentation and stereo matching. Through extensive experiments on the KITTI, vKITTI2, and Cityscapes datasets, along with both qualitative and quantitative analyses, we validate the effectiveness of our developed strategies and loss function. Our approach demonstrates superior performance compared to prior arts, with a notable increase in mean intersection over union by over 9\%.

\end{abstract}

\def\abstractname{Note to Practitioners}
\begin{abstract}
TiCoSS is a robust and effective joint learning framework that can simultaneously tackle semantic segmentation and stereo matching tasks. This work aims to improve semantic segmentation performance by exploring the potential complementarity and tightening the coupling between these two tasks. In the future, we plan to further improve the efficiency of the framework, so as to enable its real-time performance on resource-constrained hardware.
\end{abstract}


\section{Introduction}
\label{sec.intro}
\subsection{Background}

\IEEEPARstart{V}{ISUAL} environment perception serves as a fundamental and front-end module in robotic systems \cite{ciftcioglu2006studies}. Semantic segmentation and stereo matching are two key visual environment perception tasks \cite{wei2024method}. The former, akin to the ventral stream in our brain, provides a pixel-level understanding of the scene \cite{fan2020sneroadseg}, while the latter, akin to the dorsal stream in our brain, mimics human binocular vision to acquire depth information \cite{chuangweiTIP}, which is crucial for 3D geometry reconstruction. These two tasks collaborate to deliver both contextual and geometric information, resulting in a comprehensive scene understanding that significantly enhances the capabilities of robotic systems~\cite{10412168}.

Nevertheless, previous studies {\cite{depthmatch,xie2021segformer,lipson2021raftstereo}} address these two tasks with separate networks, which limits their potential to share informative contextual and geometric information \cite{zhan2019dsnet}. For instance, stereo matching networks can occasionally produce ambiguous disparity estimations, particularly in texture-less and occluded regions \cite{tase1}. Semantic segmentation can provide pixel-level scene understanding results, which help resolve such ambiguities and ultimately lead to more reliable disparity estimations \cite{wu2019semantic}. In addition, semantic segmentation networks often struggle to distinguish clear object boundaries, particularly in complex driving scenarios, due to the lack of spatial geometric information \cite{whentits,tase2}. A common solution for improved semantic segmentation performance is to employ feature-fusion networks equipped with duplex encoders to extract heterogeneous features from RGB-X data \cite{10504607},  where ``X'' provides spatial geometric information, such as the depth images generated from LiDAR point clouds \cite{10231003} and surface normal maps obtained through depth-to-normal translation \cite{fan2020sneroadseg}. However, the availability and quality of ``X" significantly influence the overall performance of semantic segmentation and can potentially limit the practical deployment of feature-fusion networks \cite{10412168}.
 
Therefore, in recent years, the simultaneous learning, deployment, and inference of both tasks have become a mainstream \cite{Zhang2018DispSegNetLS,yang2018segstereo,dovesi2020real}, because a unified joint learning framework can process contextual and geometric information more comprehensively. It also enables end-to-end training of the entire system, capable of tackling the challenges posed by both tasks~\cite{sgroadseg,sgroadseg+}. Consequently, this joint learning approach can enhance the overall performance of both semantic segmentation and stereo matching, and outperform models trained separately for each task \cite{10412168}. 

\subsection{Existing Challenges and Motivation}
The performance of a feature-fusion semantic segmentation network is heavily influenced by the employed strategy for heterogeneous feature fusion \cite{tase3,ruifanTIP}. Currently, the
bottleneck lies in the simplistic and indiscriminate fusion
of heterogeneous features, which often causes conflicting learning representations and erroneous segmentation results \cite{10643711}. Taking the state-of-the-art (SoTA) joint learning method S$^3$M-Net \cite{10412168} as an example, its adopted
feature fusion strategy essentially performs an element-wise summation between contextual and geometric feature maps at each stage. These feature maps are then directly fed into subsequent layers without filtering out irrelevant information, leading to a loose coupling between the semantic segmentation and stereo matching tasks within the encoder. Furthermore, as the network goes deeper, such an indiscriminate feature fusion strategy tends to diminish the proportion of informative geometric features in the decoder’s input \cite{feng2024sne}, potentially leading to unsatisfactory semantic segmentation performance.

Additionally, due to the vanishing gradient problem, existing joint learning frameworks often suffer from slow convergence during training. A common solution is to employ the deep supervision (DS) strategies that incorporate additional pathways to achieve gradient propagation \cite{wang2021sne}. Nonetheless, existing DS strategies employed in feature-fusion networks typically overlook the potential interactions between the main and side auxiliary classifiers, which can limit the overall semantic segmentation performance within our joint learning framework.

In the loss function aspect, previous joint learning frameworks, such as SegStereo \cite{yang2018segstereo} and SSNet \cite{jia2024ssnet}, typically compute the losses for two tasks independently, and supervise the entire training process by simply minimizing the weighted sum of these losses. Such loss function fails to leverage the potential complementarity between the two tasks at the output level, which also results in a loose coupling. 

Prior arts, such as S$^3$M-Net \cite{10412168} and DSNet \cite{zhan2019dsnet}, primarily focus on introducing a joint learning framework that performs semantic segmentation and stereo matching simultaneously. However, exploring the potential complementarity and tightening the coupling between these two tasks have received relatively limited attention in this research area and warrant further investigation.

\subsection{Contributions}

To address the aforementioned limitations, we introduce \textbf{Ti}ghtly-\textbf{Co}upled Semantic \textbf{S}egmentation and \textbf{S}tereo Matching Network (\textbf{TiCoSS}), an end-to-end joint learning approach that focuses primarily on improving the coupling between stereo matching and semantic segmentation, which has not been emphasized in previous relevant studies. Our proposed TiCoSS introduces three new techniques: (1) a \textbf{t}ightly-coupled, \textbf{g}ated feature \textbf{f}usion (TGF) strategy, which utilizes a series of selective inheritance gates (SIGs) to propagate useful contextual and geometric information from the preceding layer to the current layer, resulting in a tightly-coupled encoder; (2) a \textbf{h}ierarchical \textbf{d}eep \textbf{s}upervision (HDS) strategy that uses the fused feature maps with the highest resolution to guide deep supervision throughout each branch, as these features contain the most abundant local spatial details; (3) a novel coupling tightening (CT) loss, consisting of a widely used stereo matching loss presented in the study \cite{lipson2021raftstereo}, the semantic consistency-guided (SCG) loss introduced in the study \cite{10412168}, a disparity inconsistency-aware (DIA) loss that leverages disparity estimation results to help distinguish clearer object boundaries, and a deep supervision consistency constraint (DSCC) loss which employs the Kullback-Leibler (KL) divergence to improve prediction consistency across outputs from all deep supervision branches. 
These contributions collectively advance S$^3$M-Net, and results in TiCoSS, a new, powerful, and tightly-coupled joint learning framework that simultaneously performs robust and accurate semantic segmentation and stereo matching tasks. Extensive experiments conducted on the vKITTI2 \cite{cabon2020vkitti2}, KITTI 2015~\cite{menze2015kitti}, and Cityscapes \cite{cityscapes} datasets unequivocally demonstrate the effectiveness of the aforementioned contributions and the superior performance of TiCoSS over other SoTA approaches.

In summary, the main contributions of this article include: 
\begin{itemize}
    \item The TGF strategy, which propagates useful contextual and geometric information from the preceding layer to the current layer, enabling more effective feature fusion for semantic segmentation;
    \item The HDS strategy, which uses the fused features with the richest local spatial details to guide deep supervision across each branch;
    \item The DIA loss and the DSCC loss that tighten the coupling between the two tasks, thereby further improving the semantic segmentation performance. 
\end{itemize}

\subsection{Article Structure}
The remainder of this article is organized as follows: 
Sect. \ref{sec.related_works} reviews related prior arts.
Sect. \ref{sec.methodology} introduces our proposed TiCoSS.
Sect. \ref{sec.experiments} compares our network with other SoTA approaches and presents the ablation studies.
Finally, we conclude this article and provide recommendations for feature work in Sect. \ref{sec.conclusion}.

\section{Literature Review}
\label{sec.related_works}

\subsection{Semantic Segmentation}
\label{sec.semantic_segmentation}
Semantic segmentation has been a long-standing challenge in the fields of computer vision and robotics over the past decade \cite{ciftcioglu2006studies}. SoTA networks generally fall into two groups: (1) single-modal networks (with a single encoder) and (2) feature-fusion networks (with multiple encoders) \cite{fan2020sneroadseg}. Early efforts primarily focused on encoder-decoder architectures for pixel-level classification. Representative examples include the DeepLab series \cite{chen2018deeplabv3plus}, as well as Transformer-based networks \cite{strudel2021segmenter, xie2021segformer, cheng2022mask2former}. The encoder extracts hierarchical, contextual feature maps from input images, while the decoder generates segmentation maps by upsampling and combining feature maps from different encoder layers. Nonetheless, these networks are limited in effectively combining heterogeneous features extracted from different sources (or modalities) of visual information, which makes it challenging to produce accurate segmentation results in scenarios with poor lighting and illumination conditions \cite{fan2020sneroadseg}. This has led researchers to focus on feature-fusion networks that can effectively fuse heterogeneous features extracted from multiple sources (or modalities) of visual information. This problem is often referred to as ``RGB-X semantic segmentation'', where ``X'' represents the additional modality (or source) of visual information, in addition to the RGB images. The most representative feature-fusion networks include convolutional neural network (CNN)-based ones, such as RTFNet \cite{sun2019RTFNet} and the SNE-RoadSeg series~\cite{fan2020sneroadseg, wang2021sne, feng2024sne}, as well as Transformer-based ones, such as OFF-Net \cite{min2022orfd}, RoadFormer \cite{10504607}, and DFormer \cite{dformer}. In this article, we design TiCoSS based on S$^3$M-Net, with a special emphasis on exploring more effective solutions for tighter coupling between semantic segmentation and stereo matching.

\subsection{Stereo Matching}
\label{sec.stereo_matching}
Owing to recent advancements in deep learning techniques, end-to-end deep stereo matching networks \cite{chang2018psmnet, cheng2020leastereo, lipson2021raftstereo, li2022crestereo} have dramatically outperformed traditional explicit programming-based stereo matching algorithms. PSMNet \cite{chang2018psmnet} introduces a spatial pyramid to capture multi-scale information and employs a series of 3D convolutional layers to aggregate both local and global contexts for cost computation. 
To address the high computational cost of 3D convolutions, researchers have sought ways to balance efficiency and accuracy in stereo matching. LEA-Stereo \cite{cheng2020leastereo}, for instance, introduces the first neural architecture search (NAS) framework to stereo matching, enabling automated architecture optimization. RAFT-Stereo \cite{lipson2021raftstereo}, a rectified stereo matching approach, developed based on RAFT \cite{teed2020raft}, uses a series of gated recurrent units to iteratively refine correlation features and improve disparity estimation accuracy. CRE-Stereo \cite{li2022crestereo} further advances this approach by introducing an adaptive group-wise correlation layer to mitigate the impact of rectification errors in stereo images, resulting in more accurate disparity estimation results. In this article, we primarily focus on improving semantic segmentation performance, and therefore, adopt the stereo matching approach used in S$^3$M-Net.

\subsection{Simultaneous Semantic Segmentation and Stereo Matching}
\label{sec.joint_learning}
Existing joint learning frameworks that simultaneously address these two tasks mainly focus on improving disparity accuracy by leveraging semantic information \cite{yang2018segstereo, Zhang2018DispSegNetLS, wu2019semantic, dovesi2020real, zhan2019dsnet}. However, discussions on utilizing disparity information to enhance semantic segmentation performance at the feature level for joint learning remain limited, except for the aforementioned ``RGB-X semantic segmentation''. These prior arts often require extensive annotated training data or involve complex training strategies for joint learning. For example, SegStereo \cite{yang2018segstereo} necessitates an initial unsupervised training phase on the large-scale Cityscapes \cite{cityscapes} dataset, followed by a subsequent supervised fine-tuning on the smaller KITTI \cite{geiger2012kitti, menze2015kitti} datasets. Similarly, the approaches introduced in \cite{wu2019semantic, dovesi2020real, chen2020sgnet} require pre-training their spatial branches for stereo matching on the large-scale SceneFlow \cite{mayer2016sceneflow} dataset before fine-tuning both semantic and spatial branches on the KITTI~\cite{ menze2015kitti} dataset. DSNet \cite{zhan2019dsnet} adopts a different joint learning strategy by alternating the training of the semantic segmentation and stereo matching networks, with parameters of each network being frozen during the training of the other. 
Nevertheless, leveraging both contextual and geometric information can be challenging, as the shared features between these two tasks are not learned in an end-to-end manner. DispSegNet \cite{Zhang2018DispSegNetLS} utilizes an embedding learned from the semantic segmentation branch to refine the initial disparity estimations. RTS$^2$Net \cite{dovesi2020real} relies on coarse-to-fine estimations in a multi-stage fashion for accurate disparity estimation. However, both methods only achieve limited improvement in semantic segmentation since they fail to leverage informative spatial details to enhance semantic segmentation performance. SSNet~\cite{jia2024ssnet} employs a single encoder to extract shareable features for both tasks. However, as demonstrated in the study \cite{vitas}, such shareable features may not be suitable for both dense prediction and geometric vision tasks.
S$^3$M-Net \cite{10412168}, MENet \cite{whentits}, and SG-RoadSeg \cite{sgroadseg} use two separate encoding branches to accomplish these two tasks simultaneously, but the weak coupling between these branches limits the integration of contextual and geometric information. In contrast to the aforementioned approaches, our proposed TiCoSS uses a tightly-coupled joint learning framework that effectively leverages both contextual and geometric information. Moreover, TiCoSS is trained in an end-to-end manner and is capable of learning accurate and robust semantic segmentation and stereo matching tasks simultaneously, even with limited training data. 
\section{Methodology}
\label{sec.methodology}

\subsection{Framework Overview}
\label{sec.framework_overview}
\begin{figure*}[t!]
	\centering
	\includegraphics[width=0.99\textwidth]{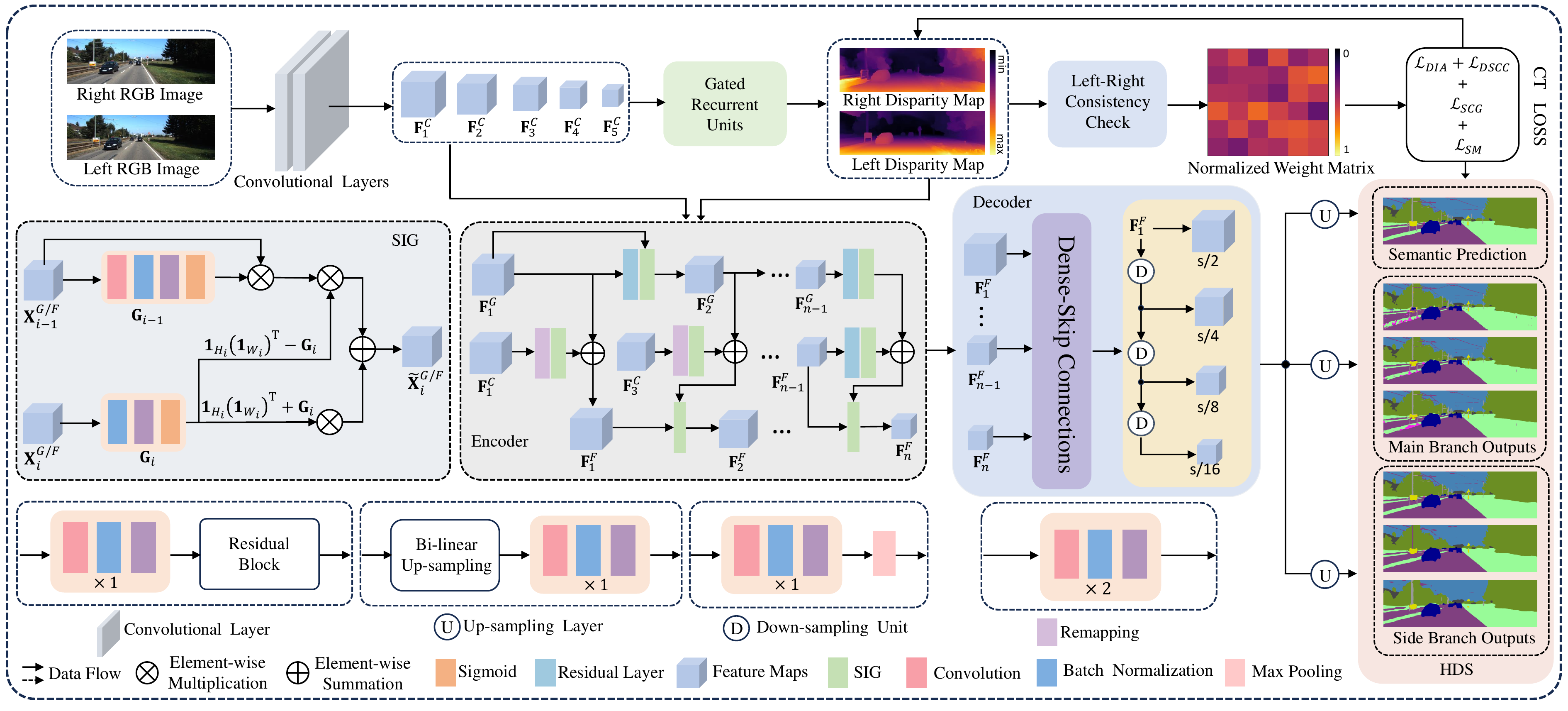}
	\caption{The architecture of our proposed TiCoSS for end-to-end joint learning of semantic segmentation and stereo matching.}
	\label{fig.pipeline}
\end{figure*}

The architecture of our proposed TiCoSS is illustrated in Fig. \ref{fig.pipeline}, containing three major technical contributions:
\begin{enumerate}[label=(\arabic*)]
    \item A novel duplex, tightly-coupled encoder designed to selectively extract and fuse heterogeneous features, namely contextual features from RGB images and geometric features from disparity maps.
    \item A novel HDS strategy that leverages fused features with the richest local spatial details to guide deep supervision across each branch (auxiliary classifier).
    \item A CT loss that supervises the entire joint learning process and further tightens the coupling between semantic segmentation and stereo matching.
\end{enumerate}

\subsection{Tightly-coupled Gated Fusion Strategy}
\label{sec.tgf}
S$^3$M-Net \cite{10412168} proposes an effective joint learning framework to simultaneously perform semantic segmentation and stereo matching. Despite achieving impressive results, these two tasks are loosely coupled. It merely employs the feature fusion strategy proposed in SNE-RoadSeg \cite{fan2020sneroadseg}, where the geometric features extracted from disparity maps are indiscriminately fused into the contextual features extracted from RGB images via simplistic element-wise summation. The fused heterogeneous features are then treated as preceding contextual features and fed into subsequent layers without selective processing, which can potentially mislead the semantic segmentation task. This is primarily because the deeper geometric features contain irrelevant semantic information, and as the network goes deeper, the proportion of contextual features in the decoder's input tends to diminish \cite{feng2024sne}. 

Our TGF strategy is, therefore, designed to overcome this limitation by selectively complementing contextual features with informative geometric features, resulting in a tightly-coupled duplex encoder. The core of our TGF strategy is the SIGs, developed based on Gated Fully Fusion (GFF) \cite{li2020gated}, which fuse features from multiple scales using gates that control the propagation of useful information. This enables the features at each scale to be enhanced by both deeper, semantically stronger features and shallower, spatially richer features, significantly reducing noises during feature fusion. Nonetheless, GFF is primarily regarded as a late fusion strategy \cite{early}, as it performs feature fusion at the decision layer and requires multi-scale features to be generated prior to processing. Additionally, GFF focuses solely on fusing features extracted from RGB images across multiple scales and is not well-suited to fuse heterogeneous features which are extracted and fused progressively. In contrast, our proposed TGF strategy  performs intermediate feature fusion during the encoding stage, enabling more interactions between heterogeneous features. Specifically, it utilizes a series of SIGs (see Fig. \ref{fig.pipeline}) to selectively inherit useful information in $\mathbf{X}_{i-1}^{G,F}$ from the previous layer into $\mathbf{X}_{i}^{G,F}$, the features at the current layer, where $i\in[1,n]\cap\mathbb{Z}$ denotes the layer number, and the superscripts `$G$' as well as `$F$' represent `geometric' and `fused' features\footnote{It is worth noting here that we use $\mathbf{X}_{i}^{F}$ instead of $\mathbf{X}_{i}^{C}$, where the superscript `$C$' denotes `contextual' features, primarily because the branch processing RGB images progressively fuses the geometric features extracted from disparity maps to obtain fused features.}, respectively. Our SIG outputs $\mathbf{\tilde{X}}_{i}^{G,F}$ selectively inherit feature maps at the $i$-th layer using the following expression:
\begin{align}
\mathbf{\tilde{X}}_{i}^{G,F} &=\Omega_{i} \left( \mathbf{X}_{i-1}^{G,F}, \mathbf{X}_{i}^{G,F} \right) = \left(\mathbf{1}_{H_i}(\mathbf{1}_{W_i})^\top+\mathbf{G}_{i}\right) \odot \mathbf{X}_{i}^{G,F} \nonumber \\
&+ \left(\mathbf{1}_{H_i}(\mathbf{1}_{W_i})^\top-\mathbf{G}_{i}\right) \odot \left [ \mathbf{G}_{i-1} \odot \mathcal{R}(\mathbf{X}_{i-1}^{G,F}) \right],
\label{eq.1}
\end{align} 
where $\Omega_i$ represents the SIG operation at the $i$-th layer, $\mathbf{1}_k$ denotes a column vector of ones, $\mathbf{G}_i \in [0,1]^{H_i \times W_i}$ represents a gate map that controls feature propagation, $\odot$ denotes the element-wise multiplication broadcasting in the channel dimension, and $\mathcal{R}$ represents the remapping operation, as detailed in the study \cite{10412168}.

Based on our proposed TGF strategy, the heterogeneous feature extraction and fusion process in our duplex encoder can be formulated as follows:
\begin{align}
&\mathbf{F}_{i}^{G}=\left\{
\begin{array}{ll}
\mathcal{E}_{i}^{G}(\mathbf{D}^L), & i=1\\
& \\
\Omega_{i}^{G}(\mathbf{F}_{i-1}^{G}\,,\,\mathcal{E}_i^G(\mathbf{F}_{i-1}^G)), & 1<i\le n 
\end{array}
\right.
\\
\text{and} \notag
\\
&\mathbf{F}_{i}^{F}=\left\{
\begin{array}{ll}
\mathcal{E}_{i}^{F}(\mathbf{F}_i^C)\:\oplus \:\mathbf{F}_i^G, & i=1\\
& \\
\Omega_{i}^{F}(\mathbf{F}_{i-1}^{F}\,,\,\mathcal{R}(\mathbf{F}_{2i-1}^C))\:\oplus\: \mathbf{F}_i^G, & 1<i\le \frac{n+1}{2} \\
& \\
\Omega_{i}^{F}(\mathbf{F}_{i-1}^{F}\,,\,\mathcal{E}_i^F(\mathbf{F}_{i-1}^F))\:\oplus\: \mathbf{F}_i^G, & \frac{n+1}{2}<i\le n
\end{array},
\right. 
\end{align}
where $\mathbf{D}^L$ denotes the estimated disparity map, $\mathbf{F}^{C}_i$, $\mathbf{F}^{G}_i$, and $\mathbf{F}^{F}_i$ respectively represent the extracted contextual feature maps in the left view, geometric feature maps, and fused feature maps at the $i$-th layer, $\mathcal{E}^G_i$ and $\mathcal{E}^F_i$ denote the geometric and fused features encoding operations at the $i$-th layer, respectively, and $\oplus$ represents the element-wise summation. Considering that the shallower features between the two tasks have similar numbers of channels, we make the contextual feature maps of the first three layers share weights with the feature maps extracted from the stereo matching network in our practical implementation. 
Our proposed TGF strategy selectively propagates useful information to subsequent layers, reducing indiscriminate heterogeneous feature fusion which can severely mislead semantic segmentation as the network goes deeper, thus achieving tighter coupling between these two relevant perception tasks. Further theoretical analyses of the TGF strategy are provided in the supplement.

\subsection{Hierarchical Deep Supervision}
\label{sec.hds}
{After tightening the coupling between semantic segmentation and stereo matching during the feature encoding stage, we turn our focus towards the feature decoding process.} We first revisit the deep supervision strategies employed in SNE-RoadSeg+ \cite{wang2021sne} and UNet 3+ \cite{Huang2020UNet3A}. The former applies deep supervision to the decoded features with the highest resolution, whereas the latter achieves this goal on the deepest decoded features at each resolution. Despite the effectiveness of these two prior approaches in addressing challenges such as vanishing gradients and slow model convergence, the deep supervision strategies employed by them are not comprehensive (the former emphasizes enhancing local details, while the latter focuses on improving consistency across multi-scale segmentation predictions). Ideally, they should be used in conjunction to complement each other for improved results. Therefore, our proposed HDS strategy combines the strengths of both methods and demonstrates superior performance compared to each individually. \\

\indent A straightforward way to combine the strengths of these two methods is to apply deep supervision strategies simultaneously to both the main and side branches, enabling the network to leverage features from shallow layers (containing rich local details) and deep layers (being semantically strong). However, this method can not fully exploit the potential complementarity between the main and side auxiliary classifiers, since the decoded features are exclusively derived from adjacently-connected ones. Additionally, as discussed in Sect. \ref{sec.tgf}, decoded features in deep layers also lack informative spatial details that are present in shallow layers, limiting the performance of DS strategies in our multi-task learning framework. Thus, to improve the interactions among auxiliary classifiers and provide local spatial details for decoded features in deep layers, we utilize the decoded, fused feature maps $\mathbf{F}_{1}^{F}$ (containing rich, fine-grained local spatial details that are essential for semantic segmentation) at the highest resolution to guide the feature decoding process in the side branches. Specifically, for the $l$-th auxiliary classifier within the side branch, we utilize a feature dynamic alignment (FDA) block, which is composed of $l$ downsampling units to progressively align channel dimensions and spatial resolutions between a pair of features at different layers. Each downsampling unit comprises a $3\times3$ convolutional layer with a stride of 2, followed by a batch normalization (BN) layer and a rectified linear unit (ReLU) activation layer. Compared to the simple max-pooling operation which may disrupt the original feature representation, our FDA achieves a smoother feature alignment. The feature maps obtained by downsampling $\mathbf{F}_{1}^{F}$ are then concatenated with the deepest decoded features at the corresponding layers. This downsampling output not only guides the decoding process but also serves as the input for the subsequent downsampling unit, thereby preserving fine-grained local details to the greatest extent.
Since the outputs of the side branches are obtained by directly upsampling the deepest features at each layer, this strategy does not significantly impact training efficiency and memory usage. Moreover, the auxiliary classifiers of the main and side branches collaboratively provide additional pathways for gradients to propagate more efficiently from the output layers to their corresponding layers, thereby accelerating the convergence of our model.

\subsection{Coupling Tightening Loss for Multi-Task Joint Learning}
\label{sec.ct-loss}
Compared to the tasks focused solely on either semantic segmentation or stereo matching, the loss function employed in our joint learning framework aims to supervise both tasks simultaneously \cite{10412168}. Our proposed CT loss function is expressed as follows:
\begin{equation}
    \mathcal{L}_{CT} = {\alpha\mathcal{L}_{DIA} + \beta\mathcal{L}_{DSCC}} + {\mathcal{L}_{SCG} + \mathcal{L}_{SM}}.
    \label{eq.ct}
\end{equation}
where the DIA loss $\mathcal{L}_{DIA}$ measures disparity inconsistency, the DSCC loss $\mathcal{L}_{DSCC}$ measures the segmentation consistency across outputs from all deep supervision branches, the SCG loss $\mathcal{L}_{SCG}$ denotes the semantic consistency-guided (SCG) loss proposed in the study \cite{10412168}, $\mathcal{L}_{SM}$, a prevalently used loss function to supervise the training of the stereo matching network, is detailed in the study \cite{10412168}, and the weight factors $\alpha$ and $\beta$ are determined through extensive ablation studies, as detailed in Sect. \ref{sec.exp_ablation}.

\subsubsection{Disparity Inconsistency-Aware Loss}
\label{sec.dia_loss}
In the previous study \cite{10412168}, the developed SCG loss focuses mainly on enforcing semantic consistency to reduce segmentation errors caused by occlusions, neglecting the fact that occlusions can also lead to inconsistent disparity estimations, which are unfortunately under-explored when designing the loss. Thus, we further incorporate the disparity inconsistency in it to form a more advanced loss function to train our TiCoSS.
Specifically, we define a weight matrix $\mathbf{W} \in \mathbb{R} ^{H\times W}$, drawing from the concept of left-right consistency check in stereo matching, where
\begin{equation}
    \mathbf{W}(\mathbf{p})=\, \mathbf{D}^L(\mathbf{p})-\mathbf{D}^R(\mathbf{p}-(\mathbf{D}^L(\mathbf{p};0)))
    \label{eq.weight_matrix}
\end{equation}
denotes the weight at the given pixel $\mathbf{p}$, and $\mathbf{D}^L$ as well as $\mathbf{D}^R$ represent the left and right disparity maps, respectively. $\mathbf{W}^{N}\in \mathbb{R} ^{H\times W}$, a normalized weight matrix, is then yielded, where 
\begin{equation}
    \mathbf{W}^{N}(\mathbf{p})= \frac{1}{1+e^{-\left | \mathbf{W}(\mathbf{p}) \right |}}\in[0,1].
    \label{eq.w_n}
\end{equation}
A higher normalized weight corresponds to a greater inconsistency between a given pair of left and right disparities, indicating the need for more attention during the training of a semantic segmentation model. Our proposed DIA loss is, therefore, formulated as follows:
\begin{equation}
    \mathcal{L}_{DIA}=\sum_{i=1}^{n} \left \{ -\frac{1}{N}\sum_{j=1}^{N}\sum_{k=1}^{C}\left [ \mathbf{W}^N(\mathbf{p})\,y_k(\boldsymbol{p})\,\mathrm{log}(\hat{y}_k(\boldsymbol{p}) ) \right ]  \right \},
    \label{eq.dia}
\end{equation}
where $n$ denotes the number of deep supervision branches, $N$ represents the total number of pixels, $C$ denotes the total number of classes, $\hat{y}_{k}(\boldsymbol{p})$ represents the predicted probability of pixel $\boldsymbol{p}$ belonging to class $k$, and $y_{k}(\boldsymbol{p})$ denotes the ground-truth label for $\boldsymbol{p}$ in class $k$.

\subsubsection{Deep Supervision Consistency Constraint Loss}
\label{sec.dscc_loss}
In prior studies \cite{wang2021sne,Huang2020UNet3A}, the relationship among prediction maps generated by different auxiliary classifiers was not considered, which may lead to semantic inconsistencies across scales. To address this issue, we draw inspiration from the study \cite{li2020dynamic} to design the following DSCC loss, which utilizes the KL divergence to measure the prediction differences across scales:
\begin{equation}
\mathcal{L}_{DSCC} = \sum_{r=1}^{L} \sum_{\substack{s=1 \\
s \neq r}}^{L}\left [ -\frac{1}{N}\sum_{i=1}^{N}\hat{y}_k^r(\boldsymbol{p}) \log\frac{\hat{y}_k^r(\boldsymbol{p})}{\hat{y}_k^s(\boldsymbol{p})} \right ] ,
\label{eq.dscc}
\end{equation}
where \textit{L} denotes the total number of auxiliary classifiers.  Further theoretical analyses of the DSCC loss are provided in the supplement.

\begin{figure*}[t!]
	\centering
	\includegraphics[width=0.99\textwidth]{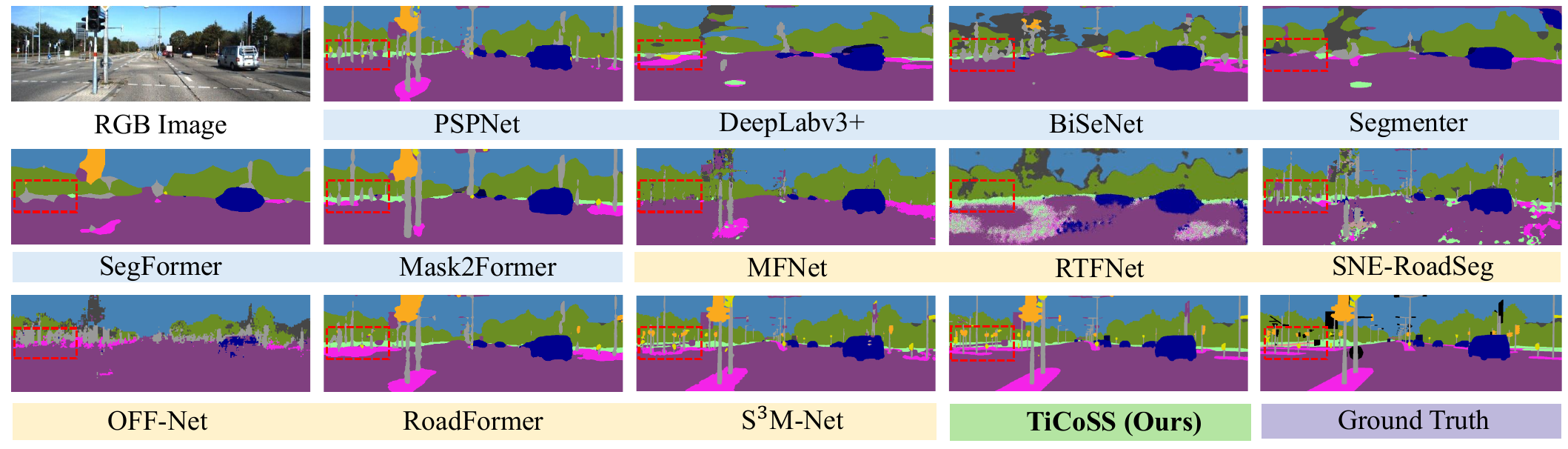}
	\caption{Qualitative experimental results achieved by the SoTA semantic segmentation networks on the KITTI 2015 \cite{menze2015kitti} dataset.}
	\label{fig.seg_kitti}
\end{figure*}

\begin{figure*}[t!]
	\centering
	\includegraphics[width=0.99\textwidth]{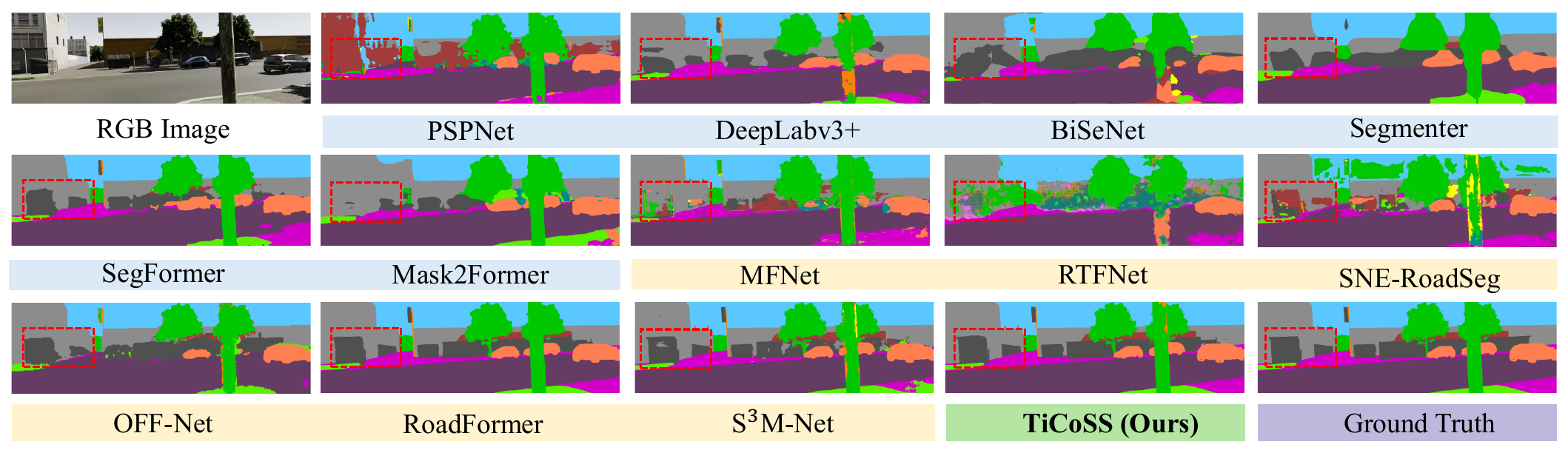}
	\caption{Qualitative experimental results achieved by the SoTA semantic segmentation networks on the vKITTI2 \cite{cabon2020vkitti2} dataset.} 
	\label{fig.seg_vkitti}
\end{figure*}

\begin{figure*}[t!]
	\centering
	\includegraphics[width=0.99\textwidth]{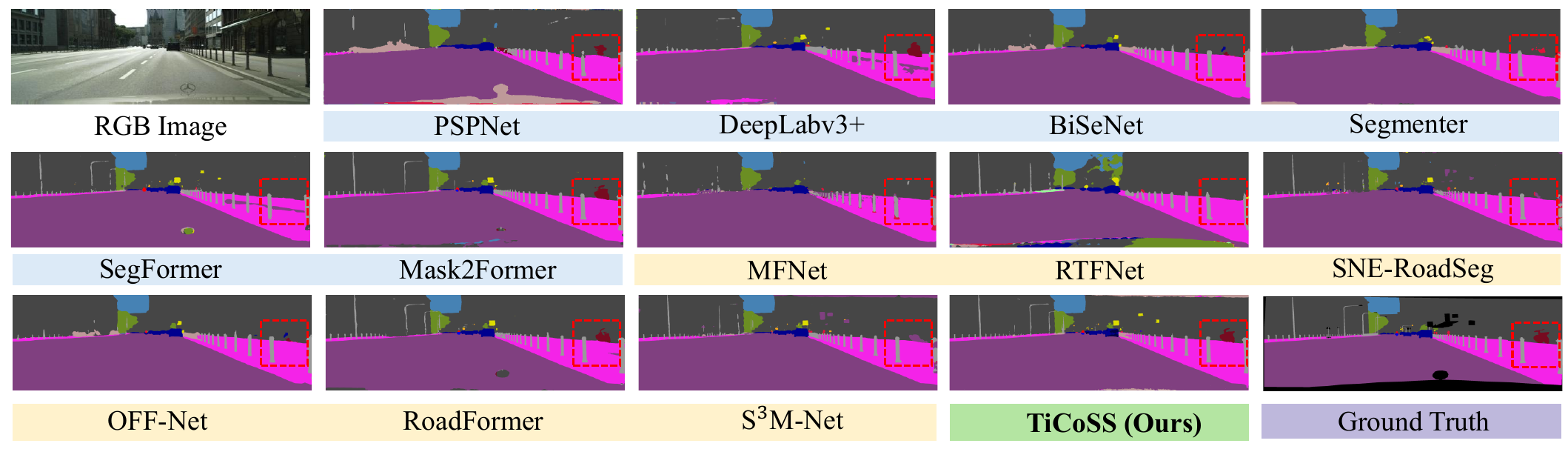}
	\caption{Qualitative experimental results achieved by the SoTA semantic segmentation networks on the Cityscapes \cite{cityscapes} dataset.} 
	\label{fig.seg_city}
\end{figure*}

\section{Experiments}
\label{sec.experiments}
In this article, we conduct extensive experiments to evaluate the performance of our introduced TiCoSS both quantitatively and qualitatively. The following subsections detail the used datasets, experimental setup, evaluation metrics, and the comprehensive evaluation of our proposed method.

\subsection{Datasets}
\label{sec.datasets}

Since our network training requires both semantic and disparity annotations, we utilize the following three public datasets to conduct the experiments:
\begin{itemize}
\item 
The vKITTI2 \cite{cabon2020vkitti2} dataset contains virtual replicas (providing 15 semantic classes) of five sequences from the KITTI dataset. Dense ground-truth disparity maps are obtained through depth rendering using a virtual engine. Following the study \cite{10412168}, we employ 700 stereo image pairs, along with their semantic and disparity annotations, to evaluate the effectiveness and robustness of our proposed TiCoSS, where 500 pairs are utilized for model training and the remaining 200 pairs are used for model validation.

\item The KITTI 2015 \cite{menze2015kitti} dataset contains 400 stereo image pairs captured in real-world driving scenarios. Half of these pairs have ground truth annotations, while the remaining half do not. This dataset provides 19 semantic classes (consistent with those in the Cityscapes \cite{cityscapes} dataset). Sparse ground-truth disparity maps are obtained using a Velodyne HDL-64E LiDAR. In our experiments, we split the data with a 7:3 ratio for training and testing, respectively.

\item The Cityscapes \cite{cityscapes} dataset is a widely
used urban scene understanding dataset, containing 2,975 stereo images for model training and 500 stereo images for model validation, with well-annotated semantic annotations. It is noteworthy that the corresponding depth information is obtained using ViTAStereo \cite{vitas}, since depth ground truth is unavailable.
\end{itemize}

\subsection{Experimental Setup}
\label{sec.experimental_setup}
Our experiments are conducted on an NVIDIA RTX 3090 GPU with a batch size of 1. We set the maximum disparity search range to 192 pixels. All images are cropped to 512 $\times$ 256 pixels before being processed by the network. We utilize the AdamW optimizer for model training, with epsilon and weight decay set to $10^{-8}$ and $10^{-5}$, respectively.
The initial learning rate is set to $2\times 10^{-4}$. Training lasts for 100,000 iterations on the vKITTI2 dataset, 20,000 iterations on the KITTI 2015 dataset, and 50,000 iterations on the Cityscapes dataset. Standard data augmentation techniques are applied to improve model robustness.

\subsection{Evaluation Metrics}
\label{sec.evaluation_metrics}

Following the study \cite{10412168}, we quantify the performance of semantic segmentation using seven metrics: (1) accuracy (Acc), (2) mean accuracy (mAcc), (3) precision (Pre), (4) recall (Rec), (5) mean F1-score (mFSc), (6) mean intersection over union (mIoU), and (7) frequency-weighted intersection over union (fwIoU). We calculate each of these metrics on a per-image basis before averaging them across the entire dataset. Moreover, we quantify the performance of stereo matching using two metrics: (1) average end-point error (EPE) and (2) percentage of error pixels (PEP) at tolerance levels of 1.0 and 3.0 pixels, respectively.

\begin{table*}[!htb]
        \settablefont
	\centering
	\caption{
        Quantitative comparisons of SoTA semantic segmentation networks on the KITTI 2015 \cite{menze2015kitti} dataset. The best results are shown in bold font. The symbol $\uparrow$ indicates that a higher value corresponds to better performance. 
	}
\label{tab.seg_kitti}
	\setlength{\tabcolsep}{1.5mm}
	\begin{tabular}
	    {L{2.0cm}|C{1.4cm}|C{1.6cm} |C{1.25cm} C{1.25cm} C{1.25cm} C{1.25cm} C{1.25cm} C{1.25cm}}
		\toprule
		  Networks & Publication & \multicolumn{1}{c|}{Type} & Acc (\%) $\uparrow$ & mAcc (\%) $\uparrow$ & mIoU (\%) $\uparrow$  & Pre (\%) $\uparrow$ & Rec (\%) $\uparrow$ & mFSc (\%) $\uparrow$ \\
		\hline
            \hline
		DeepLabv3+ \cite{chen2018deeplabv3plus} & ECCV'18 &\multirow{5}{*}{Single-Modal} & 82.33 & 50.15 & 42.79  & 83.85 & 87.18 & 84.59 \\
		BiSeNet V2 \cite{yu2021bisenetv2} & IJCV'21 & & 73.68 & 41.66 & 32.71  & 68.35 & 81.79 & 72.37 \\
		Segmenter \cite{strudel2021segmenter} & ICCV'21 & & 84.53 & 50.77 & 43.63 & 82.99 & 87.15 & 84.41 \\
		SegFormer \cite{xie2021segformer} & NeurIPS'21 & & 88.28 & 59.23 & 51.39  & 87.15 & 90.85 & 88.46 \\
		Mask2Former \cite{cheng2022mask2former} & CVPR'22 & & 84.35 & 54.33 & 45.87 & 84.74 & 89.12 & 85.92 \\
		\hline
		RTFNet \cite{sun2019RTFNet} & RAL'19 &\multirow{6}{*}{Feature-Fusion} & 71.61 & 39.26 & 30.35 & 69.52 & 85.16 & 74.28 \\
		SNE-RoadSeg \cite{fan2020sneroadseg} & ECCV'20 & & 79.46 & 51.91 & 41.56 & 81.45 & 87.05 & 82.91 \\
        OFF-Net \cite{min2022orfd} & ICRA'22 & & 75.84 & 40.13 & 33.13 & 77.48 & 72.19 & 70.62 \\
		RoadFormer \cite{10504607} & TIV'24 & & 90.05 & 62.34 & 55.13 & 91.65 & 91.39 & 91.11 \\
        RoadFormer+ \cite{10643711} & TIV'24 & & 91.35 & 66.29 & 57.69 & 91.24 & 92.07 & 91.47 \\
        \textcolor{black}{DFormer \cite{dformer}} & \textcolor{black} {ICLR'24} &  &\textcolor{black}{90.59} &\textcolor{black}{69.01}  &\textcolor{black}{58.18}  &\textcolor{black}{91.20}  &\textcolor{black}{93.58}  &\textcolor{black}{91.96} \\ 
        \hline
        {DispSegNet \cite{Zhang2018DispSegNetLS}} &  {RA-L'18} &\multirow{5}{*}{Joint Learning} & {88.03} &  {62.54} &  {53.44}  &  {88.22} & {92.53} &  {89.68}  \\
         {DSNet\cite{zhan2019dsnet}} &  {ICRA'19}  && {89.48} &  {64.44} &  {52.83} &  {89.41} & {93.25} &  {90.73} \\
         \textcolor{black} {SG-RoadSeg \cite{sgroadseg}} & \textcolor{black} {ICRA'24} & &\textcolor{black} {86.51}  &\textcolor{black} {61.10}  &\textcolor{black} {52.02}  &\textcolor{black} {88.46}  &\textcolor{black} {80.10}  &\textcolor{black}  {84.96} \\
         S$^3$M-Net \cite{10412168} & TIV'24 & & 90.66 & 65.90 & 57.80 & 90.85 & 93.55 & 91.80 \\
        \textbf{TiCoSS} (\textbf{Ours}) & TASE'25 & &\textbf{91.90}  &\textbf{71.97} &\textbf{63.63}  &\textbf{92.43}  &\textbf{94.10} &\textbf{92.90}  \\
    	\bottomrule
		\end{tabular}
\end{table*}

\begin{table*}[!htb]
        \settablefont
	\centering
	\caption{
        Quantitative comparisons of SoTA semantic segmentation networks on the vKITTI2 \cite{cabon2020vkitti2} dataset. The best results are shown in bold font. The symbol $\uparrow$ indicates that a higher value corresponds to better performance.
	}
\label{tab.seg_vkitti}
	\setlength{\tabcolsep}{1.5mm}
	\begin{tabular}
	    {L{2.0cm}|C{1.4cm}|C{1.6cm} |C{1.25cm} C{1.25cm} C{1.25cm} C{1.25cm} C{1.25cm} C{1.25cm}}
		\toprule
		  Networks & Publication & \multicolumn{1}{c|}{Type} & Acc (\%) $\uparrow$ & mAcc (\%) $\uparrow$ & mIoU (\%) $\uparrow$  & Pre (\%) $\uparrow$ & Rec (\%) $\uparrow$ & mFSc (\%) $\uparrow$ \\
		\hline
            \hline
		DeepLabv3+ \cite{chen2018deeplabv3plus} & ECCV'18 &\multirow{5}{*}{Single-Modal}& 92.19 & 63.15 & 56.90  & 89.00 & 92.71 & 90.16 \\
		BiSeNet V2 \cite{yu2021bisenetv2} & IJCV'21 && 81.77 & 51.07 & 44.45  & 83.23 & 82.19 & 80.67 \\
		Segmenter \cite{strudel2021segmenter} & ICCV'21 && 90.39 & 60.33 & 52.99 & 88.05 & 87.89 & 87.70 \\
		SegFormer \cite{xie2021segformer} & NeurIPS'21 && 94.75 & 70.56 & 64.98 & 93.57 & 93.62 & 93.46 \\
		Mask2Former \cite{cheng2022mask2former} & CVPR'22 && 89.29 & 64.58 & 57.14 & 90.75 & 87.23 & 87.19 \\
		\hline
		RTFNet \cite{sun2019RTFNet} & RAL'19 & \multirow{6}{*}{Feature-Fusion} & 85.22 & 49.47 & 42.59 & 83.74 & 89.17 & 84.41 \\
		SNE-RoadSeg \cite{fan2020sneroadseg} & ECCV'20 && 83.64 & 60.85 & 52.56 & 83.44 & 81.66 & 77.77 \\
        OFF-Net \cite{min2022orfd} & ICRA'22 && 90.84 & 61.51 & 55.27 & 89.24 & 86.71 & 86.15\\
		RoadFormer \cite{10504607} & TIV'24 && 97.54 & 86.58 & 80.83 & 96.99 & 96.86 & 96.91 \\
        RoadFormer+ \cite{10643711} & TIV'24 && 98.45 & 89.79 & 84.03 & 96.36 & 96.94 & 97.77 \\
        \textcolor{black}{DFormer \cite{dformer}} & \textcolor{black}{ICLR'24} &  & \textcolor{black} {97.79} & \textcolor{black}{89.06} &\textcolor{black}{85.54} &\textcolor{black}{96.41}  &\textcolor{black}{94.67}  &\textcolor{black}{95.58} \\ 
        \hline
         {DispSegNet \cite{Zhang2018DispSegNetLS}} &  {RA-L'18} &\multirow{5}{*}{Joint Learning} & {97.15} &  {82.63} &  {80.24} &  {97.39} & {97.66} &  {97.46}\\
        {DSNet\cite{zhan2019dsnet}} &  {ICRA'19} && {95.71} &  {78.71} &  {77.47} &  {97.20} & {96.66} &  {96.81} \\
        \textcolor{black}{SG-RoadSeg \cite{sgroadseg}} &\textcolor{black} {ICRA'24} & &\textcolor{black} {95.17}  &\textcolor{black} {85.57}  &\textcolor{black} {80.91}  &\textcolor{black} {86.10}  &\textcolor{black} {88.50}  &\textcolor{black} {86.31}\\
        S$^3$M-Net \cite{10412168} & TIV'24 && 98.32 & 88.24 & 84.18 & 98.37 & 98.28 & 98.31 \\
        \textbf{TiCoSS} (\textbf{Ours}) & TASE'25 & &\textbf{98.69}  &\textbf{91.66}  & \textbf{88.46} &\textbf{98.55}  &\textbf{98.67} &\textbf{98.57}  \\
    	\bottomrule
		\end{tabular}
\end{table*}
    
\begin{table*}[t!]
        \settablefont
    \color{black}
	\centering
	\caption{
        Quantitative comparisons of SoTA semantic segmentation networks on the Cityscapes \cite{cityscapes} dataset. The best results are shown in bold font. The symbol $\uparrow$ indicates that a higher value corresponds to better performance. Values marked with ``-'' denote that the corresponding metrics were not reported in the original paper.
        }
\label{tab.seg_cityscapes}
	\setlength{\tabcolsep}{1.5mm}
    \vspace{2mm}
	\begin{tabular}
	    {L{2.0cm}|C{1.6cm}|C{1.8cm} |C{1.25cm} C{1.25cm} C{1.25cm} C{1.25cm} C{1.25cm} C{1.25cm}}
		\toprule
		  Networks & Publication & \multicolumn{1}{c|}{Type} & Acc (\%) $\uparrow$ & mAcc (\%) $\uparrow$ & mIoU (\%) $\uparrow$  & Pre (\%) $\uparrow$ & Rec (\%) $\uparrow$ & mFSc (\%) $\uparrow$\\
		\hline
            \hline
        DeepLabv3+ \cite{chen2018deeplabv3plus} & ECCV'18 &\multirow{5}{*}{Single-Modal}  &84.11 &62.75 &50.15 &71.12 &57.34 &59.96  \\
		BiSeNet V2 \cite{yu2021bisenetv2} & IJCV'21 & &85.61 &61.16 &49.65 &71.00 &57.61 &62.50  \\
		Segmenter \cite{strudel2021segmenter} & ICCV'21 & &90.19 &79.87 &62.76 &76.90 &80.99 &77.19  \\
        SegFormer \cite{xie2021segformer} & NeurIPS'21 &  &88.12 &73.79 &62.48 &75.50 &78.81 &71.02 \\
        Mask2Former \cite{cheng2022mask2former} & CVPR'22 & & 87.97 & 78.80 & 64.29 & 75.01 & 78.26 & 74.71 \\
		\hline
        RTFNet \cite{sun2019RTFNet} & RAL'19 &\multirow{6}{*}{Feature-Fusion} &88.41 &67.16 &50.19 &69.41 &70.07 &67.40 \\
        SNE-RoadSeg \cite{fan2020sneroadseg} & ECCV'20 &  &86.60 &75.47 &63.01 &70.56 &74.19 &75.58  \\
        OFF-Net \cite{min2022orfd} & ICRA'22 & & 76.88 & 60.03 & 43.11 & 70.68 & 71.01 & 69.65 \\
        RoadFormer \cite{10504607} & TIV'24 &  &87.11 &75.49 &62.20 &74.49 &76.84 &75.50 \\
        RoadFormer+ \cite{10643711} & TIV'24 & & 90.40 & 78.86 & 64.18 & 77.19 & 80.06 & 76.60 \\
        {DFormer} \cite{dformer} & ICLR'24 &  &\textbf{91.37} &80.16 &65.59 &76.60  &79.65  &77.41 \\ 
        \hline
         {DispSegNet \cite{Zhang2018DispSegNetLS}} &  {RA-L'18} &\multirow{6}{*}{Joint Learning} &84.07 &62.18 &56.10 &70.00 &60.92 &61.16 \\
         {DSNet \cite{zhan2019dsnet}} &  {ICRA'19}  & &84.01 &62.25 &52.79 &69.56 &63.74 &61.26 \\
         {MENet \cite{whentits}} &  {TITS'21}  &  &93.39 &69.53 &61.50 & - & - & - \\
         {SG-RoadSeg \cite{sgroadseg}} & ICRA'24 & &85.80 &66.95 &54.18 &74.10 &67.92 &70.10 \\
        {S$^3$M-Net \cite{10412168}} & TIV'24 &  &88.47  &77.30  &62.59  &75.20  &77.48  &76.30 \\ 
        \textbf{TiCoSS (Ours)} & TASE'25 &  &90.70  &\textbf{81.76}  &\textbf{68.36}  &\textbf{78.43}  &\textbf{81.76} &\textbf{79.74}\\
    	\bottomrule
		\end{tabular}
\end{table*}

\begin{table}
        \settablefont
	\centering
	\caption{
        Quantitative comparisons between TiCoSS and the baseline S$^3$M-Net, evaluated using the mmsegmentation framework. The best results are shown in bold font. The symbol $\uparrow$ indicates that a higher value corresponds to better performance.
	}
\label{tab.mmseg_comparison}
	\setlength{\tabcolsep}{1.5mm}
	\begin{tabular}
	    {C{1.8cm}| C{1.65cm} C{1.65cm} C{1.65cm}}
		\toprule
            \multirow{2}{*}{Methods}  & \multicolumn{3}{c}{mIoU (\%) $\uparrow$}\\
            \cline{2-4}
            &{vKITTI2} &{KITTI 2015} &{Cityscapes}\\
		\hline    \hline
             S$^3$M-Net \cite{10412168}  &84.00 &41.54 &55.40\\
            \hline
            \textbf{TiCoSS} (\textbf{Ours})      &\textbf{87.94}  & \textbf{47.66} & \textbf{62.16}\\
            \bottomrule
		\end{tabular}
\end{table}

\begin{table*}[ht]
        \fontsize{6.8}{11.5}\selectfont
	\centering
	\caption{
        Comparisons of SoTA stereo matching network on the KITTI 2015 \cite{menze2015kitti} and vKITTI2 \cite{cabon2020vkitti2} datasets. The symbol $\downarrow$ indicates that a lower value corresponds to better performance. The best results are shown in bold font.
	}
\label{tab.disp_comparisons}
	\setlength{\tabcolsep}{1.5mm}
	\begin{threeparttable}
	\begin{tabular}
	    {L{2.4cm}| C{1.6cm}|C{1.6cm} C{1.45cm} C{1.45cm}| C{1.6cm} C{1.45cm} C{1.45cm} }
        \toprule 
           \multicolumn{1}{c|}{\multirow{3}{*}{Networks}} & \multirow{3}{*}{Publications}  & \multicolumn{3}{c}{vKITTI2 \cite{cabon2020vkitti2}} & \multicolumn{3}{c}{KITTI 2015 \cite{menze2015kitti}} \\

             &&\multirow{2}{*}{EPE (pixels) $\downarrow$} & \multicolumn{2}{c|}{PEP (\%) $\downarrow$} &\multirow{2}{*}{EPE (pixels) $\downarrow$} & \multicolumn{2}{c}{PEP (\%) $\downarrow$} \\
            \cline{4-5} \cline{7-8}
            && & $>1$ pixel & $>3$ pixels & & $>1$ pixel & $>3$ pixels\\
            \hline \hline
		PSMNet \cite{chang2018psmnet} &CVPR'18       & 0.68 & 10.31 & 3.77  & 0.74 & 16.12 & 2.61\\ 
		LEA-Stereo \cite{cheng2020leastereo} &NeurIPS'20  & 0.83 & 13.33 & 4.84  & 0.83 & 18.67 & 3.22 \\
		RAFT-Stereo \cite{lipson2021raftstereo} &3DV'21    & 0.40 & 5.88 & 2.67 & 0.60 & 10.78 & 1.96\\
        CRE-Stereo \cite{li2022crestereo} &CVPR'22   & 0.63 & 10.35 & 3.90  & 0.92 & 19.68 & 3.35 \\
        ACVNet \cite{xu2022acvnet} &CVPR'22           & 0.61 & 9.41 & 3.45  & 0.68 & 13.93 & 2.10\\
        PCW-Net \cite{shen2022pcwnet} &ECCV'22        & 0.63 & 9.45 & 3.49  & 0.70 & 14.81 & 2.43 \\
        IGEV-Stereo \cite{xu2023iterative} &CVPR'23   & 0.47 & 7.15 & 3.09  & 0.62 & 12.15 & 1.99 \\
        {DispSegNet \cite{Zhang2018DispSegNetLS}} &  {RA-L'18}  &  {0.50} &  {6.37} & {2.91} & {0.81} &  {14.47} &  {2.69}
       \\
        {DSNet\cite{zhan2019dsnet}}  & {ICRA'19}  &  {0.64} &  {8.42} & {3.82} & {0.73} &  {15.10} &  {2.92}\\
        S$^3$M-Net \cite{10412168} & TIV'24              & 0.38 & 5.56 & \textbf{2.55} & 0.55 & \textbf{10.02} & 1.62 \\
        \textbf{TiCoSS (Ours)} &TASE'25  &\textbf{0.34} &\textbf{5.43} &2.58 &\textbf{0.54} &10.39 &\textbf{1.60}\\
        
	\Xhline{1.5pt}
		\end{tabular}
	\end{threeparttable}
\end{table*}

\subsection{Comparison with State-of-The-Art Methods}
\label{sec.SoTA_comparisions}
\subsubsection{Semantic Segmentation Performance}
\label{sec.semntic_performance}
The qualitative experimental results on the KITTI, vKITTI2, and Cityscapes datasets are presented in Figs. \ref{fig.seg_kitti}, \ref{fig.seg_vkitti}, and \ref{fig.seg_city}, respectively, while the quantitative experimental results on these three datasets are given in Tables \ref{tab.seg_kitti}, \ref{tab.seg_vkitti}, and \ref{tab.seg_cityscapes}, respectively.
We also compare the semantic segmentation performance of TiCoSS and the baseline S$^3$M-Net using the mmsegmentation framework\footnote{The mmsegmentation framework is available at \url{https://github.com/open-mmlab/mmsegmentation}.}, with the results provided in Table \ref{tab.mmseg_comparison}.

These results suggest that TiCoSS outperforms all other SoTA single-modal and feature-fusion networks (including both CNN-based and Transformer-based methods) across most evaluation metrics on these three datasets. Specifically, compared with S$^3$M-Net, the SoTA joint learning method, TiCoSS demonstrates substantial improvements on the KITTI dataset, achieving increases of $9.68\%$ in mAcc, $10.57\%$ in mIoU, $2.71\%$ in fwIoU, and $1.20\%$ in mFSc, respectively. 

Similarly, on the vKITTI2 dataset, it outperforms other networks across all evaluation metrics, with improvements of $3.88\%$ in mAcc, $5.08\%$ in mIoU, $0.62\%$ in fwIoU, and $0.26\%$ in mFSc, respectively. 
On the Cityscapes dataset, it outperforms other networks in most evaluation metrics, with improvements of $1.99\%$ in mAcc, $4.22\%$ in mIoU, $1.60\%$ in Pre, $2.12\%$ in Rec, and $3.00\%$ in mFSc.
Particularly, as observed in Figs. \ref{fig.seg_kitti} and \ref{fig.seg_vkitti}, TiCoSS achieves more accurate predictions on distant regions as well as object boundaries and is capable of providing more fine-grained semantic segmentation details compared to S$^3$M-Net. 

We attribute these improvements to the tighter coupling between the two tasks within our joint learning framework, which effectively leverages the informative geometric information extracted from our predicted disparity maps. This enhances the integration of contextual and geometric features through our proposed TGF strategy, further improving the ability to handle heterogeneous features. 

\subsubsection{Stereo Matching Performance}
\label{sec.exp_disp}
\begin{figure*}[!htb]
	\centering
	\includegraphics[width=0.99\textwidth]{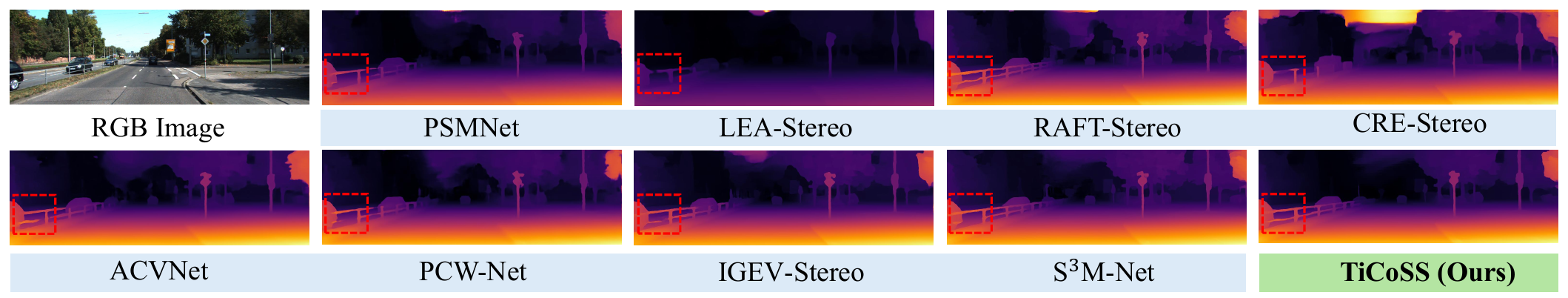}
	\caption{Qualitative experimental results achieved by the SoTA stereo matching networks  on the KITTI 2015 \cite{menze2015kitti} dataset.}
	\label{fig.disp_kitti}
\end{figure*}

\begin{figure*}[!htb]
	\centering
	\includegraphics[width=0.99\textwidth]{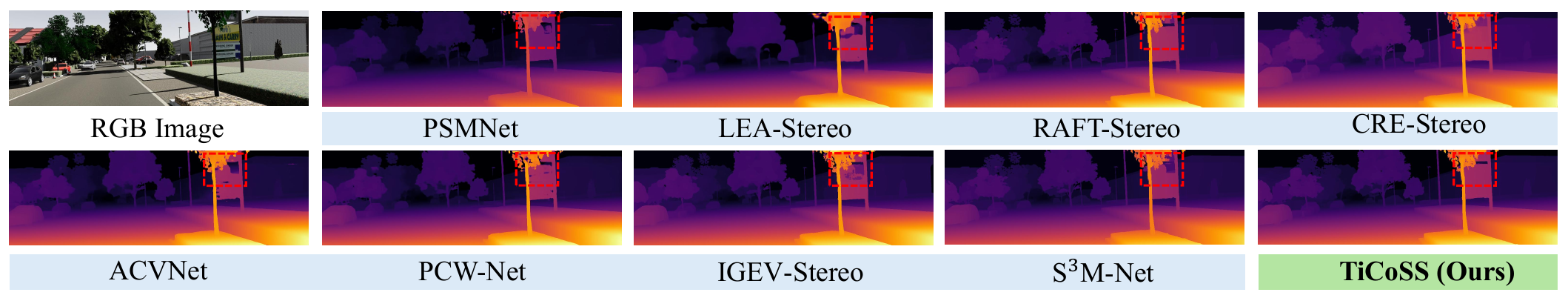}
	\caption{Qualitative experimental results achieved by the SoTA stereo matching networks on the vKITTI2 \cite{cabon2020vkitti2} dataset.}
	\label{fig.disp_vkitti}
\end{figure*}

The qualitative experimental results on the KITTI and vKITTI2 datasets are illustrated in Figs. \ref{fig.disp_kitti} and \ref{fig.disp_vkitti}, respectively, while the quantitative experimental results on these two datasets are provided in Table \ref{tab.disp_comparisons}. Since the primary focus of this study is to improve semantic segmentation performance, the stereo matching performance of TiCoSS is slightly better than that of S$^3$M-Net. Specifically, compared to S$^3$M-Net, TiCoSS demonstrates improvements of $3.64\%$ in EPE and $2.47\%$ in PEP 3.0 on the KITTI dataset. Additionally, on the vKITTI2 dataset, it achieves improvements of $5.26\%$ in EPE and $1.26\%$ in PEP 1.0. 
These improvements can be attributed to the tighter coupling between the two tasks, resulting in more comprehensive geometric features with informative contextual information compared to S$^3$M-Net. Additionally, by minimizing the CT loss, our model focuses more on areas with inconsistent disparities and achieves improved performance in occluded regions, as depicted in Fig. \ref{fig.disp_vkitti}.

\subsection{Ablation Studies}
\label{sec.exp_ablation}

\begin{table}
        \settablefont
	\centering
	\caption{
        Quantitative comparisons between our proposed TGF strategy and two SoTA feature fusion strategies, ASFF and GFF, on the KITTI 2015\cite{menze2015kitti} dataset. ``\textbf{Baseline}": S$^3$M-Net w/o SCG loss \cite{10412168}. The symbol $\uparrow$ indicates that a higher value corresponds to better performance. The best results are shown in bold font.
	}
\label{tab.TGF_comparison}
	\setlength{\tabcolsep}{1.4mm}
	\begin{tabular}
        {L{2.4 cm}| C{1.15cm} | C{1.15cm}| C{1.3cm} C{1.3cm}}
		\toprule
            \multirow{2}{*}{Feature Fusion Strategies}
            & Disparity
            & RGB
            & \multirow{2}{*}{mIoU (\%) $\uparrow$}
            & \multirow{2}{*}{fwIoU (\%) $\uparrow$}
            \\
            & Branch & Branch && \\
		\hline \hline
            Baseline           &&   &54.33 &83.44 \\
            \hline
            \multirow{3}{*}{Baseline + ASFF \cite{Liu2019LearningSF}}
                    &\checkmark &                  &54.72  &83.10  \\
            
            &&\checkmark                 & 54.16    &78.25    \\
            &\checkmark &\checkmark   & 55.92    &  83.88         \\   
            \hline 
            \multirow{3}{*}{Baseline + GFF \cite{li2020gated}}
                    &\checkmark&                  &57.13    &84.33             \\ 
            &&\checkmark                & 57.66       &83.69       \\
            &\checkmark &\checkmark       &58.03 &84.39       \\
            \hline 
            \multirow{3}{*}{Baseline + TGF (\textbf{Ours})}
                    &\checkmark&                 &55.80 &84.30        \\
            &&\checkmark                 & 58.44       &82.87      \\
             &\checkmark &\checkmark       &\textbf{59.06}       & \textbf{84.72}   \\
            \bottomrule
		\end{tabular}
\end{table}

\subsubsection{Heterogeneous Feature Fusion Strategy}
\label{tgf_abaltion} 
Extensive experiments are conducted on the KITTI 2015\cite{menze2015kitti} dataset to validate the effectiveness of our proposed TGF strategy, compared with two other SoTA feature fusion strategies: adaptive spatial feature fusion (ASFF) \cite{Liu2019LearningSF} and GFF \cite{li2020gated}. As presented in Table \ref{tab.TGF_comparison}, TGF outperforms both ASFF and GFF, with an
increase in mIoU by up to $7.93\%$, and an increase in fwIoU by $8.26\%$. These compelling results demonstrate the capability of TGF for effective feature extraction and for selective feature propagation. Additionally, we demonstrate the rationality of employing feature fusion strategies within each encoding branch. It is evident that all models perform best when feature fusion is adopted in both branches, demonstrating the necessity to selectively extract and fuse heterogeneous features within both encoding branches.

\subsubsection{Deep Supervision Strategy}
\label{hds_selction}
We first compare the performance of our proposed HDS strategy with single-resolution deep supervision (SDS) \cite{wang2021sne}, full-scale deep supervision (FDS) \cite{Huang2020UNet3A}, and the combination of them (referred to SDS + FDS). As presented in Table \ref{tab.HDS_comparison}, our proposed HDS achieves the SoTA performance across both evaluation metrics, with improvements of up to $5.59\%$ in mIoU and $2.02\%$ in fwIoU. These compelling results can be attributed to the improved interactions among auxiliary classifiers, leveraging fused features with the richest local spatial details in the main branch to guide deep supervision across side branches. Additionally, it is noteworthy that the straightforward combination of SDS and FDS results in a marginal improvement compared to using FDS alone. 
Additionally, SDS is adopted only at the highest resolution, where features have a low number of channels, resulting in almost no additional memory usage. Therefore, we incorporate SDS into our HDS. Furthermore, we conduct an additional ablation study to evaluate the performance of HDS when utilizing guidance features in different layers and of various types. As shown in Table \ref{tab.HDS_feature_guide}, the fused features at the shallowest layer result in the best performance compared to another two. We attribute this superior performance to the rich, fine-grained local spatial details that $\mathbf{F}_{1}^{F}$ contains, which are essential for semantic segmentation.

\subsubsection{Loss Function}
\label{loss_selction}
\begin{figure}[htbp]
    \centering
    \includegraphics[width=0.47 \textwidth]{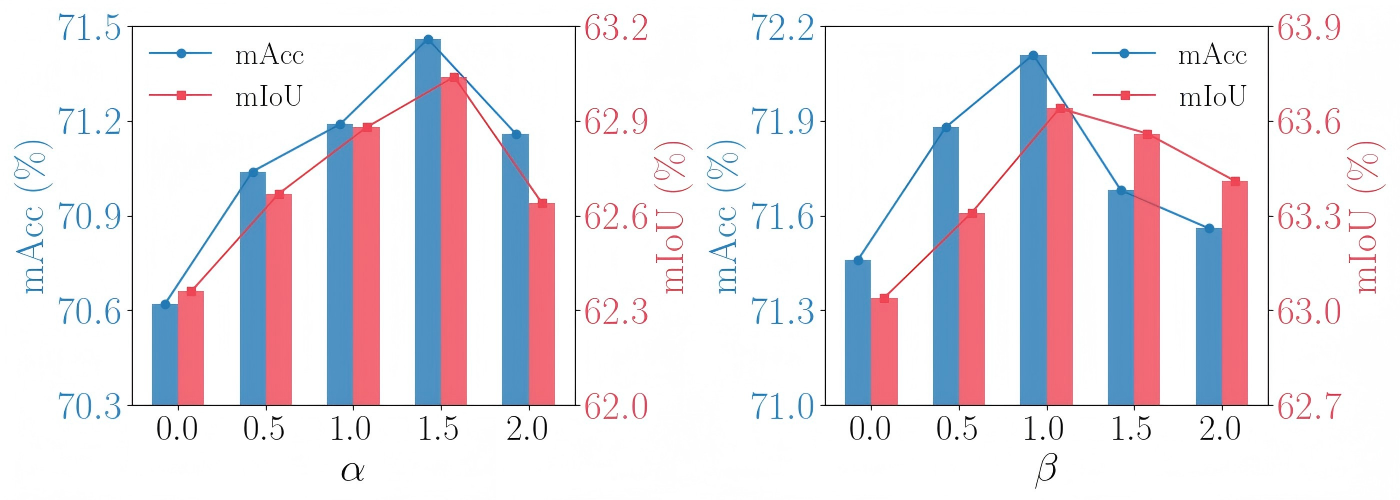}
    \caption{
     The selection of  hyperparameters $\alpha$ and $\beta$ within the CT loss on the KITTI 2015 \cite{menze2015kitti} dataset.
    }
    \label{fig.alpha_beta}
\end{figure}

 We systematically incorporate each element into the CT loss to assess its influence on the overall performance. Experimental results presented in Table \ref{tab.loss_abaltion} validate the effectiveness of each component of our proposed CT loss. It is evident that the DIA and DSCC losses make significant contributions to semantic segmentation. Notably, when the entire joint learning framework is trained by minimizing the CT loss, TiCoSS achieves the best performance on the KITTI dataset, attaining a maximum mIoU of $63.63\%$.
 Additionally, to maximize the effectiveness of our proposed loss function, we first conduct an ablation study on the selection of loss weight $\alpha$ in (\ref{eq.dia}). Fig. \ref{fig.alpha_beta} shows the mAcc and mIoU values with respect to different $\alpha$ within the range of $0.0$ to $2.0$. It can be obviously observed that when $\alpha=1.5$, TiCoSS achieves the best overall performance for both evaluation metrics. Following the selection of $\alpha$, we conduct another ablation study to determine $\beta$ in (\ref{eq.dscc}). Fig. \ref{fig.alpha_beta} demonstrates that $\beta = 1.0$ is the best choice. Further weight tuning is possible, but it should be approached cautiously, especially when dealing with limited data to avoid over-fitting.

\subsubsection{All Contributions}
\label{overall_performance}
We explore the rationality of each contribution adopted in our TiCoSS. As presented in Table \ref{tab.overall_comparison}, we sequentially incorporate each novel component to assess its impact on the overall performance. Please note that our CT loss is based on HDS strategy. Therefore, no ablation study needs to be conducted solely with the CT loss. The results demonstrate that each component employed in our model contributes to an improvement in the overall performance, and the network achieves its peak performance when all novel contributions (TGF, HDS, and CT loss) are leveraged, demonstrating the effectiveness of our design.

\begin{table}
        \settablefont
	\centering
	\caption{
       Ablation study on our HDS strategy on the KITTI 2015 \cite{menze2015kitti} dataset. ``\textbf{Baseline}":  S$^3$M-Net \cite{10412168} enhanced by our TGF strategy. The symbol $\uparrow$ indicates that a higher value corresponds to better performance. The best results are shown in bold font.
	}
\label{tab.HDS_comparison}
	\setlength{\tabcolsep}{1.5mm}
	\begin{tabular}
	    {L{3.5cm}| C{1.45cm} C{1.45cm}}
		\toprule
            Methods  & mIoU (\%) $\uparrow$ & fwIoU (\%) $\uparrow$ \\
		\hline    \hline
            Baseline                        &59.06      & 84.72        \\
            \hline
            Baseline + SDS \cite{wang2021sne}                     & 60.01    &84.79  \\
            \hline
            Baseline + FDS  \cite{Huang2020UNet3A}                  &60.62      & 85.20    \\
            \hline
            Baseline + SDS + FDS   &60.86 &85.59 \\
            \hline
            Baseline + HDS (\textbf{Ours})  &\textbf{62.36}       &\textbf{86.33}  \\
            \bottomrule
		\end{tabular}
\end{table}

\begin{table}
        \settablefont
	\centering
	\caption{
        Ablation study on the selection of the guidance features employed in our HDS strategy on the KITTI 2015 \cite{menze2015kitti} dataset. ``\textbf{Feature Layer}": the layer index of the guidance feature, with options of 1, 2, and 3. ``\textbf{GF}": Geometric Features, ``\textbf{CF}": Contextual Features, ``\textbf{FF}": Fused Features, obtained by performing an element-wise summation of the geometric and contextual features. The symbol $\uparrow$ indicates that a higher value corresponds to better performance. The best results are shown in bold font. 
	}
\label{tab.HDS_feature_guide}
	\setlength{\tabcolsep}{1.5mm}
	\begin{tabular}
	    {C{0.55cm} C{0.55 cm} C{0.55 cm} | C{0.55 cm} C{0.55 cm} C{0.55 cm}| C{1.45cm} C{1.45cm}  } 
		\toprule
            \multicolumn{3}{c|}{Feature Layer}
            &\multicolumn{3}{c|}{Method} 
            &\multirow{2}{*}{ mIoU  (\%) $\uparrow$ }
            &\multirow{2}{*}{ fwIoU  (\%) $\uparrow$ }
            \\
            \cline{1-6}
            1 &2 &3 &GF &CF &FF & &    \\
            \hline \hline
           \checkmark & &    &\checkmark & &       &61.74 &\textbf{86.88} \\
           \checkmark & &    & &\checkmark &        &62.09 &86.48 \\
           \checkmark & &    & & &\checkmark        &\textbf{62.36}       &86.33    \\
           \hline
           
            &\checkmark &   &\checkmark & &         &61.18 &85.52\\
            &\checkmark &   & &\checkmark &         &61.12 &86.85\\
            &\checkmark &   & & &\checkmark         &62.11  &86.31\\
           \hline
           
            & &\checkmark &\checkmark & &          &56.67 &83.68 \\
            & &\checkmark & &\checkmark &          &59.05  &84.11\\
            & &\checkmark & & &\checkmark          &60.26  &85.60 \\
        \bottomrule
		\end{tabular}
\end{table}

\begin{table}
        \settablefont
	\centering
	\caption{
        Ablation study to validate the effectiveness of the three semantic segmentation losses within our CT loss on the KITTI 2015 \cite{menze2015kitti} dataset. The symbol $\uparrow$ indicates that a higher value corresponds to better performance. The best results are shown in bold font.}
\label{tab.loss_abaltion}
	\setlength{\tabcolsep}{1.5mm}
	\begin{tabular}
	    { C{1cm}  C{1cm} C{1cm}| C{1.45cm} C{1.45cm}  }
		\toprule
            DIA &DCSS &SCG 
              & mIoU (\%) $\uparrow$ & fwIoU (\%) $\uparrow$\\
		\hline \hline
             &&             	&62.36	&86.33\\
            \checkmark &&  	&63.04	&86.48\\
             &\checkmark&     &62.88	&85.98\\
            &&\checkmark         &62.75	&86.66\\
           \checkmark&\checkmark&    &63.54	&86.50\\
            \checkmark&&\checkmark   &63.15	&86.7\\
            &\checkmark&\checkmark   &62.99	&86.59\\
            \checkmark&\checkmark&\checkmark  &\textbf{63.63} &\textbf{86.68}	\\						
            \bottomrule
		\end{tabular}
\end{table}

\begin{table}
        \settablefont
	\centering
	\caption{
        Ablation study on the three contributions on the KITTI 2015 \cite{menze2015kitti} dataset. The symbol $\uparrow$ indicates that a higher value corresponds to better performance. The best results are shown in bold font.}
\label{tab.overall_comparison}
	\setlength{\tabcolsep}{1.5mm}
	\begin{tabular}
	    { C{1cm}  C{1cm} C{1cm}| C{1.4cm} C{1.4cm}  }
		\toprule
            TGF &HDS &CT loss 
               & mIoU (\%) $\uparrow$   & mFSc (\%) $\uparrow$ \\
		\hline \hline
            \checkmark &&             &59.06  &91.26   \\
             &\checkmark&             &58.49  &92.10   \\
           \checkmark&\checkmark&     &62.36  &92.74   \\
            &\checkmark&\checkmark    &61.18  &92.46   \\
            \checkmark&\checkmark&\checkmark            &\textbf{63.63}  &\textbf{92.90}  \\						
            \bottomrule
		\end{tabular}
\end{table}

\subsection{Efficiency Analysis}
Additionally, TiCoSS contains 385.05 million trainable parameters and requires 308.86 GFLOPs to process an image with a resolution of 512$\times$256 pixels. When deployed on an NVIDIA GeForce RTX 3090 GPU paired with an Intel Core i7-13700KF processor, it achieves an inference speed of 0.30 seconds per image, with a memory usage of around 5.82 GB.
We believe that the further reduction of TiCoSS’s computational complexity is essential for deployment on resource-constrained hardware.

\subsection{Additional Experiments}
 We conduct several additional experiments, as detailed in the supplement to comprehensively validate the effectiveness of TiCoSS. Specifically, we evaluate TiCoSS's robustness by conducting extensive experiments under various weather conditions and challenging scenarios. Moreover, we submit our results to the KITTI Semantics benchmark to compare TiCoSS with methods whose codes are not publicly available. These additional qualitative and quantitative experimental results further demonstrate the superior performance of TiCoSS.
\section{Conclusion and Future Work}
\label{sec.conclusion}
{This article introduced TiCoSS, a novel, high-performing, and state-of-the-art joint learning framework designed to tighten the coupling between the semantic segmentation and stereo matching tasks. We made three key contributions in this work: (1) an effective feature fusion strategy and a tightly-coupled duplex encoder, leveraging the informative spatial information to enhance the semantic segmentation task,} (2) a novel hierarchical deep supervision strategy that improves the interactions among all auxiliary classifiers, and (3) a joint learning loss that focuses on further tightening the coupling of these two tasks at the output level. The effectiveness of each contribution was validated through extensive experiments.

Despite its superior performance over existing approaches, TiCoSS still requires both semantic and disparity annotations, and collecting data with such ground truth remains a labor-intensive process. Thus, exploring semi-supervised or few-shot semantic segmentation methods, and un/self-supervised stereo matching methods is a promising avenue for our future research.
\bibliographystyle{IEEEtran}
\bibliography{refs}
\end{document}

%% file: packages.tex
\usepackage[table,xcdraw]{xcolor}
\usepackage{scalerel}
\usepackage{tikz}
\usetikzlibrary{svg.path}

\definecolor{orcidlogocol}{HTML}{A6CE39}
\tikzset{
  orcidlogo/.pic={
    \fill[orcidlogocol] svg{M256,128c0,70.7-57.3,128-128,128C57.3,256,0,198.7,0,128C0,57.3,57.3,0,128,0C198.7,0,256,57.3,256,128z};
    \fill[white] svg{M86.3,186.2H70.9V79.1h15.4v48.4V186.2z}
                 svg{M108.9,79.1h41.6c39.6,0,57,28.3,57,53.6c0,27.5-21.5,53.6-56.8,53.6h-41.8V79.1z M124.3,172.4h24.5c34.9,0,42.9-26.5,42.9-39.7c0-21.5-13.7-39.7-43.7-39.7h-23.7V172.4z}
                 svg{M88.7,56.8c0,5.5-4.5,10.1-10.1,10.1c-5.6,0-10.1-4.6-10.1-10.1c0-5.6,4.5-10.1,10.1-10.1C84.2,46.7,88.7,51.3,88.7,56.8z};
  }
}
\newcommand\orcidicon[1]{\href{https://orcid.org/#1}{\mbox{\scalerel*{
\begin{tikzpicture}[yscale=-1,transform shape]
\pic{orcidlogo};
\end{tikzpicture}
}{|}}}}

%% file: final.bbl
\begin{thebibliography}{10}
\providecommand{\url}[1]{#1}
\csname url@samestyle\endcsname
\providecommand{\newblock}{\relax}
\providecommand{\bibinfo}[2]{#2}
\providecommand{\BIBentrySTDinterwordspacing}{\spaceskip=0pt\relax}
\providecommand{\BIBentryALTinterwordstretchfactor}{4}
\providecommand{\BIBentryALTinterwordspacing}{\spaceskip=\fontdimen2\font plus
\BIBentryALTinterwordstretchfactor\fontdimen3\font minus \fontdimen4\font\relax}
\providecommand{\BIBforeignlanguage}[2]{{%
\expandafter\ifx\csname l@#1\endcsname\relax
\typeout{** WARNING: IEEEtran.bst: No hyphenation pattern has been}%
\typeout{** loaded for the language `#1'. Using the pattern for}%
\typeout{** the default language instead.}%
\else
\language=\csname l@#1\endcsname
\fi
#2}}
\providecommand{\BIBdecl}{\relax}
\BIBdecl

\bibitem{ciftcioglu2006studies}
{\"O}.~Ciftcioglu \emph{et~al.}, ``{Studies on Visual Perception for Perceptual Robotics},'' in \emph{International Conference on Robotics and Automation (ICRA)}, vol.~2.\hskip 1em plus 0.5em minus 0.4em\relax SCITEPRESS, 2006, pp. 352--359.

\bibitem{wei2024method}
F.~Wei and W.~Wang, ``{A Method for Designing the Perception Module of Autonomous Vehicles Using Stereo Depth and Semantic Segmentation},'' in \emph{2024 24th International Conference on Control, Automation and Systems (ICCAS)}.\hskip 1em plus 0.5em minus 0.4em\relax IEEE, 2024, pp. 1423--1428.

\bibitem{fan2020sneroadseg}
R.~Fan \emph{et~al.}, ``{SNE-RoadSeg: Incorporating Surface Normal Information into Semantic Segmentation for Accurate Freespace Detection},'' in \emph{Proceedings of the European Conference on Computer Vision (ECCV)}.\hskip 1em plus 0.5em minus 0.4em\relax Springer, 2020, pp. 340--356.

\bibitem{chuangweiTIP}
C.-W. Liu \emph{et~al.}, ``{These Maps Are Made by Propagation: Adapting Deep Stereo Networks to Road Scenarios With Decisive Disparity Diffusion},'' \emph{IEEE Transactions on Image Processing}, vol.~34, pp. 1516--1528, 2025.

\bibitem{10412168}
Z.~Wu \emph{et~al.}, ``{S$^3$M-Net: Joint Learning of Semantic Segmentation and Stereo Matching for Autonomous Driving},'' \emph{IEEE Transactions on Intelligent Vehicles}, vol.~9, no.~2, pp. 3940--3951, 2024.

\bibitem{depthmatch}
J.~Huang \emph{et~al.}, ``{DepthMatch: Semi-Supervised RGB-D Scene Parsing through Depth-Guided Regularization},'' \emph{IEEE Signal Processing Letters}, pp. 1--5, 2025, {DOI}:{10.1109/LSP.2025.3575640}.

\bibitem{xie2021segformer}
E.~Xie \emph{et~al.}, ``{SegFormer: Simple and Efficient Design for Semantic Segmentation with Transformers},'' \emph{Advances in Neural Information Processing Systems (NeurIPS)}, vol.~34, pp. 12\,077--12\,090, 2021.

\bibitem{lipson2021raftstereo}
Lipson \emph{et~al.}, ``{RAFT-Stereo: Multilevel Recurrent Field Transforms for Stereo Matching},'' in \emph{International Conference on 3D Vision (3DV)}.\hskip 1em plus 0.5em minus 0.4em\relax IEEE, 2021, pp. 218--227.

\bibitem{zhan2019dsnet}
W.~Zhan \emph{et~al.}, ``{DSNet: Joint Learning for Scene Segmentation and Disparity Estimation},'' in \emph{International Conference on Robotics and Automation (ICRA)}.\hskip 1em plus 0.5em minus 0.4em\relax IEEE, 2019, pp. 2946--2952.

\bibitem{tase1}
X.~Guan \emph{et~al.}, ``{Multistage Pixel-Visibility Learning With Cost Regularization for Multiview Stereo},'' \emph{IEEE Transactions on Automation Science and Engineering}, vol.~20, no.~2, pp. 751--762, 2023.

\bibitem{wu2019semantic}
Z.~Wu \emph{et~al.}, ``{Semantic Stereo Matching with Pyramid Cost Volumes},'' in \emph{Proceedings of the IEEE/CVF International Conference on Computer Vision (ICCV)}, 2019, pp. 7484--7493.

\bibitem{whentits}
X.~Zhang \emph{et~al.}, ``{When Visual Disparity Generation Meets Semantic Segmentation: A Mutual Encouragement Approach},'' \emph{IEEE Transactions on Intelligent Transportation Systems}, vol.~22, no.~3, pp. 1853--1867, {2021}.

\bibitem{tase2}
Y.~Sun \emph{et~al.}, ``{FuseSeg: Semantic Segmentation of Urban Scenes Based on RGB and Thermal Data Fusion},'' \emph{IEEE Transactions on Automation Science and Engineering}, vol.~18, no.~3, pp. 1000--1011, 2021.

\bibitem{10504607}
J.~Li \emph{et~al.}, ``{RoadFormer: Duplex Transformer for RGB-Normal Semantic Road Scene Parsing},'' \emph{IEEE Transactions on Intelligent Vehicles}, pp. 1--10, 2024, {DOI}:{10.1109/TIV.2024.3388726}.

\bibitem{10231003}
J.~Zhang \emph{et~al.}, ``{CMX: Cross-Modal Fusion for RGB-X Semantic Segmentation With Transformers},'' \emph{IEEE Transactions on Intelligent Transportation Systems}, vol.~24, no.~12, pp. 14\,679--14\,694, 2023.

\bibitem{Zhang2018DispSegNetLS}
{J. Zhang \textit{et al.}}, ``{DispSegNet: Leveraging Semantics for End-to-end Learning of Disparity Estimation from Stereo Imagery},'' \emph{IEEE Robotics and Automation Letters}, vol.~4, pp. 1162--1169, 2018.

\bibitem{yang2018segstereo}
G.~Yang \emph{et~al.}, ``{SegStereo: Exploiting Semantic Information for Disparity Estimation},'' in \emph{Proceedings of the European Conference on Computer Vision (ECCV)}, 2018, pp. 636--651.

\bibitem{dovesi2020real}
P.~L. Dovesi \emph{et~al.}, ``{Real-Time Semantic Stereo Matching},'' in \emph{2020 IEEE International Conference on Robotics and Automation (ICRA)}, 2020, pp. 10\,780--10\,787.

\bibitem{sgroadseg}
Z.~Wu \emph{et~al.}, ``{SG-RoadSeg: End-to-End Collision-Free Space Detection Sharing Encoder Representations Jointly Learned via Unsupervised Deep Stereo},'' in \emph{2024 IEEE International Conference on Robotics and Automation (ICRA)}, {2024}, pp. {11\,063--11\,069}.

\bibitem{sgroadseg+}
M.-J. Lee \emph{et~al.}, ``{SG-RoadSeg+: End-to-End Freespace Detection Upgraded at Data, Feature, and Loss Levels},'' \emph{IEEE Transactions on Instrumentation and Measurement}, pp. 1--1, 2025, {DOI}: 10.1109/TIM.2025.3579733.

\bibitem{tase3}
W.~Zhou \emph{et~al.}, ``{CMPFFNet: Cross-Modal and Progressive Feature Fusion Network for RGB-D Indoor Scene Semantic Segmentation},'' \emph{IEEE Transactions on Automation Science and Engineering}, vol.~21, no.~4, pp. 5523--5533, 2024.

\bibitem{ruifanTIP}
R.~Fan \emph{et~al.}, ``{Graph Attention Layer Evolves Semantic Segmentation for Road Pothole Detection: A Benchmark and Algorithms},'' \emph{IEEE Transactions on Image Processing}, vol.~30, pp. 8144--8154, 2021.

\bibitem{10643711}
J.~Huang \emph{et~al.}, ``{RoadFormer+: Delivering RGB-X Scene Parsing through Scale-Aware Information Decoupling and Advanced Heterogeneous Feature Fusion},'' \emph{IEEE Transactions on Intelligent Vehicles}, pp. 1--10, 2024, {DOI}:{10.1109/TIV.2024.3448251}.

\bibitem{feng2024sne}
Y.~Feng \emph{et~al.}, ``{SNE-RoadSegV2: Advancing Heterogeneous Feature Fusion and Fallibility Awareness for Freespace Detection},'' \emph{IEEE Transactions on Instrumentation and Measurement}, vol.~74, pp. 1--9, 2025.

\bibitem{wang2021sne}
H.~Wang \emph{et~al.}, ``{SNE-RoadSeg+: Rethinking Depth-Normal Translation and Deep Supervision for Freespace Detection},'' in \emph{IEEE/RSJ International Conference on Intelligent Robots and Systems (IROS)}.\hskip 1em plus 0.5em minus 0.4em\relax IEEE, 2021, pp. 1140--1145.

\bibitem{jia2024ssnet}
D.~Jia \emph{et~al.}, ``{SSNet: a Joint Learning Network for Semantic Segmentation and Disparity Estimation},'' \emph{The Visual Computer}, pp. 1--13, 2024.

\bibitem{cabon2020vkitti2}
Y.~Cabon \emph{et~al.}, ``{Virtual KITTI 2},'' 2020.

\bibitem{menze2015kitti}
M.~Menze and A.~Geiger, ``{Object Scene Flow for Autonomous Vehicles},'' in \emph{Proceedings of the IEEE/CVF Conference on Computer Vision and Pattern Recognition (CVPR)}, 2015, pp. 3061--3070.

\bibitem{cityscapes}
M.~Cordts \emph{et~al.}, ``{The Cityscapes Dataset for Semantic Urban Scene Understanding},'' in \emph{Proceedings of the IEEE/CVF Conference on Computer Vision and Pattern Recognition (CVPR)}, 2016, pp. 3213--3223.

\bibitem{chen2018deeplabv3plus}
{L.-C. Chen et al}, ``{Encoder-Decoder with Atrous Separable Convolution for Semantic Image Segmentation},'' in \emph{Proceedings of the European Conference on Computer Vision (ECCV)}, September 2018.

\bibitem{strudel2021segmenter}
R.~Strudel \emph{et~al.}, ``{SegMenter: Transformer for Semantic Segmentation},'' in \emph{Proceedings of the IEEE/CVF International Conference on Computer Vision (ICCV)}, 2021, pp. 7262--7272.

\bibitem{cheng2022mask2former}
B.~Cheng \emph{et~al.}, ``{Masked-Attention Mask Transformer for Universal Image Segmentation},'' in \emph{Proceedings of the IEEE/CVF Conference on Computer Vision and Pattern Recognition (CVPR)}, 2022, pp. 1290--1299.

\bibitem{sun2019RTFNet}
Y.~Sun \emph{et~al.}, ``{RTFNet: RGB-Thermal Fusion Network for Semantic Segmentation of Urban Scenes},'' \emph{IEEE Robotics and Automation Letters}, vol.~4, no.~3, pp. 2576--2583, 2019.

\bibitem{min2022orfd}
C.~Min \emph{et~al.}, ``{ORFD: A Dataset and Benchmark for Off-Road Freespace Detection},'' in \emph{2022 International Conference on Robotics and Automation (ICRA)}, 2022, pp. 2532--2538.

\bibitem{dformer}
B.~Yin \emph{et~al.}, ``{DFormer: Rethinking RGBD Representation Learning for Semantic Segmentation},'' in \emph{International Conference on Learning Representations (ICLR)}, 2024.

\bibitem{chang2018psmnet}
J.-R. Chang and Y.-S. Chen, ``{Pyramid Stereo Matching Network},'' in \emph{Proceedings of the IEEE/CVF Conference on Computer Vision and Pattern Recognition (CVPR)}, 2018, pp. 5410--5418.

\bibitem{cheng2020leastereo}
X.~Cheng \emph{et~al.}, ``{Hierarchical Neural Architecture Search for Deep Stereo Matching},'' \emph{Advances in Neural Information Processing Systems (NeurIPS)}, vol.~33, pp. 22\,158--22\,169, 2020.

\bibitem{li2022crestereo}
J.~Li \emph{et~al.}, ``{Practical Stereo Matching via Cascaded Recurrent Network with Adaptive Correlation},'' in \emph{Proceedings of the IEEE/CVF Conference on Computer Vision and Pattern Recognition (CVPR)}, 2022, pp. 16\,263--16\,272.

\bibitem{teed2020raft}
Teed \emph{et~al.}, ``{RAFT: Recurrent All-Pairs Field Transforms for Optical Flow},'' in \emph{Proceedings of the European Conference on Computer Vision (ECCV)}.\hskip 1em plus 0.5em minus 0.4em\relax Springer, 2020, pp. 402--419.

\bibitem{geiger2012kitti}
A.~Geiger \emph{et~al.}, ``{Are We Ready for Autonomous Driving? the KITTI Vision Benchmark Suite},'' in \emph{Proceedings of the IEEE/CVF Conference on Computer Vision and Pattern Recognition (CVPR)}.\hskip 1em plus 0.5em minus 0.4em\relax IEEE, 2012, pp. 3354--3361.

\bibitem{chen2020sgnet}
S.~Chen \emph{et~al.}, ``{SGNet: Semantics Guided Deep Stereo Matching},'' in \emph{Proceedings of the Asian Conference on Computer Vision (ACCV)}, 2020, pp. 106--122.

\bibitem{mayer2016sceneflow}
N.~Mayer \emph{et~al.}, ``{A Large Dataset to Train Convolutional Networks for Disparity, Optical Flow, and Scene Flow Estimation},'' in \emph{Proceedings of the IEEE/CVF Conference on Computer Vision and Pattern Recognition (CVPR)}, 2016, pp. 4040--4048.

\bibitem{vitas}
C.-W. Liu \emph{et~al.}, ``{Playing to Vision Foundation Model's Strengths in Stereo Matching},'' \emph{IEEE Transactions on Intelligent Vehicles}, pp. 1--12, 2024, {DOI}: 10.1109/TIV.2024.3467287.

\bibitem{li2020gated}
X.~Li \emph{et~al.}, ``{Gated Fully Fusion for Semantic Segmentation},'' in \emph{Proceedings of the AAAI Conference on Artificial Intelligence (AAAI)}, vol.~34, no.~07, 2020, pp. 11\,418--11\,425.

\bibitem{early}
Boulahia \emph{et~al.}, ``{Early, Intermediate and Late Fusion Strategies for Robust Deep Learning-based Multimodal Action Recognition},'' \emph{Machine Vision and Applications}, vol.~32, no.~6, p. 121, 2021.

\bibitem{Huang2020UNet3A}
H.~Huang \emph{et~al.}, ``{UNet 3+: A Full-Scale Connected UNet for Medical Image Segmentation},'' \emph{ICASSP 2020 - 2020 IEEE International Conference on Acoustics, Speech and Signal Processing (ICASSP)}, pp. 1055--1059, 2020.

\bibitem{li2020dynamic}
D.~Li and Q.~Chen, ``{Dynamic Hierarchical Mimicking Towards Consistent Optimization Objectives},'' in \emph{Proceedings of the IEEE/CVF Conference on Computer Vision and Pattern Recognition (CVPR)}, 2020, pp. 7642--7651.

\bibitem{yu2021bisenetv2}
C.~Yu \emph{et~al.}, ``{BiSeNet V2: Bilateral Network with Guided Aggregation for Real-Time Semantic Segmentation},'' \emph{International Journal of Computer Vision}, vol. 129, pp. 3051--3068, 2021.

\bibitem{xu2022acvnet}
G.~Xu \emph{et~al.}, ``{Attention Concatenation Volume for Accurate and Efficient Stereo Matching},'' in \emph{Proceedings of the IEEE/CVF Conference on Computer Vision and Pattern Recognition (CVPR)}, 2022, pp. 12\,981--12\,990.

\bibitem{shen2022pcwnet}
Z.~Shen \emph{et~al.}, ``{PCW-Net: Pyramid Combination and Warping Cost Volume for Stereo Matching},'' in \emph{Proceedings of the European Conference on Computer Vision (ECCV)}.\hskip 1em plus 0.5em minus 0.4em\relax Springer, 2022, pp. 280--297.

\bibitem{xu2023iterative}
G.~Xu \emph{et~al.}, ``{Iterative Geometry Encoding Volume for Stereo Matching},'' in \emph{Proceedings of the IEEE/CVF Conference on Computer Vision and Pattern Recognition (CVPR)}, 2023, pp. 21\,919--21\,928.

\bibitem{Liu2019LearningSF}
S.~Liu \emph{et~al.}, ``{Learning Spatial Fusion for Single-Shot Object Detection},'' \emph{ArXiv}, vol. abs/1911.09516, 2019.

\end{thebibliography}
